\documentclass{article} 
\usepackage{iclr2025_conference,times}

\usepackage{wrapfig}

\usepackage{amsmath,amsfonts,bm}
\usepackage{multicol}

\def\eqref#1{equation~\ref{#1}}

\def\1{\bm{1}}

\DeclareMathAlphabet{\mathsfit}{\encodingdefault}{\sfdefault}{m}{sl}
\SetMathAlphabet{\mathsfit}{bold}{\encodingdefault}{\sfdefault}{bx}{n}

\usepackage{booktabs}      
\usepackage{multirow}      
\usepackage{amssymb}       
\usepackage{graphicx}     
\usepackage{tabularx}
\usepackage{colortbl}
\usepackage{xcolor}
\usepackage{array}
\usepackage{makecell}
\usepackage{hyperref}
\usepackage{url}
\usepackage{multicol}
\usepackage{aliascnt}
\usepackage{adjustbox}
\usepackage{tcolorbox}
\usepackage{siunitx} 
\usepackage{enumitem}
\usepackage{amsmath}  
\usepackage{longtable} 
\usepackage{subcaption}
\usepackage{calc}
\usepackage{tikz}
\usepackage{etoolbox}
\usepackage{ragged2e}
\usepackage{fvextra}
\DefineVerbatimEnvironment{AppendixPrompt}{Verbatim}{
  breaklines=true,
  breakanywhere=true,
  fontsize=\small,
  formatcom=\ttfamily,
  frame=single,
  framesep=2mm
}

\definecolor{barblue}{RGB}{49,59,135}
\definecolor{bargray}{RGB}{217,217,217}
\definecolor{textgray}{RGB}{90,90,90}

\newcommand{\best}[1]{\textbf{#1}}
\newcommand{\second}[1]{\underline{#1}}

\newcommand{\scorebar}[2]{%
\begin{tikzpicture}[baseline=(n.base)]
  \node[inner sep=0pt, outer sep=0pt, anchor=west] (n) at (0,0) {%
    \begin{tikzpicture}[x=1mm,y=1mm,baseline=(m.base)]
      \fill[bargray] (0,0) rectangle (28,2.2);
      \pgfmathsetmacro{\w}{28*#1/100}
      \fill[barblue] (0,0) rectangle (\w,2.2);
      \node[anchor=west, font=\scriptsize] (m) at (29.2,1.1) {#2};
    \end{tikzpicture}%
  };
\end{tikzpicture}%
}

\definecolor{errbarblue}{RGB}{220,37,42}   
\definecolor{errbargray}{RGB}{235,235,235} 

\newcommand{\scorebarerr}[2]{%
\begin{tikzpicture}[baseline=(n.base)]
  \node[inner sep=0pt, outer sep=0pt, anchor=west] (n) at (0,0) {%
    \begin{tikzpicture}[x=1mm,y=1mm,baseline=(m.base)]
      \fill[errbargray] (0,0) rectangle (15,2.2);
      \pgfmathsetmacro{\w}{15*#1/100}
      \fill[errbarblue] (0,0) rectangle (\w,2.2);
      \node[anchor=west, font=\scriptsize] (m) at (16.2,1.1) {#2};
    \end{tikzpicture}%
  };
\end{tikzpicture}%
}

\newcommand{\deltainline}[2]{#1 {\scriptsize (#2)}}

\definecolor{uclablue}{rgb}{0.15, 0.45, 0.68}
\hypersetup{
    breaklinks,
    citecolor=uclablue,
    colorlinks=true,
}

\newtcolorbox{AIbox}[2][]{aibox,title=#2,#1}
\tcbset{
  aibox/.style={
    top=10pt,
    colframe=black,
    colbacktitle=black,
    center
  }
}

\title{PlanningBench: Generating Scalable and Verifiable Planning Data for Evaluating and Training Large Language Models}

\author{
Ziliang Zhao$^{1,2,*}$, Zenan Xu$^{2,*}$, Shuting Wang$^1$, Hongjin Qian$^3$, Yan Lei$^2$, Minda Hu$^{2,4}$, \mbox{Zhao~Wang}$^1$, Shihan Dou$^2$, Zhicheng Dou$^{1,\dagger}$, Pluto Zhou$^{2,\dagger}$  \\
\vspace{1mm}
\textbf{$^1$Gaoling School of Artificial Intelligence, Renmin University of China} \quad
\textbf{$^2$LLM Department, Hunyuan Team, Tencent} \quad
\textbf{$^3$Beijing Academy of Artificial Intelligence} \quad
\textbf{$^4$The Chinese University of Hong Kong} \\
\vspace{1mm}
{\small $^*$Equal contribution.\quad $^\dagger$Corresponding author.}
}

\begin{document}
\maketitle
\let\oldthefootnote\thefootnote

\let\thefootnote\oldthefootnote

\begin{abstract}
Planning is a fundamental capability for large language models (LLMs) because such complex tasks require models to coordinate goals, constraints, resources, and long-term consequences into executable and verifiable solutions. Existing planning benchmarks, however, usually treat planning data as fixed collections of instances rather than controllable generation targets. This limits scenario coverage, ties difficulty to surface-level proxies rather than structural sources, and offers limited support for scalable generation, automatic verification, or planning-oriented training. We introduce \textbf{PlanningBench}, a framework for generating scalable, diverse, and verifiable planning data for both evaluation and training. PlanningBench starts from real planning scenarios and abstracts practical workflows into a structured taxonomy of more than 30 task types, subtasks, constraint families, and difficulty factors. Guided by this taxonomy, a constraint-driven synthesis pipeline instantiates self-contained planning problems with adaptive difficulty control, quality filtering, and instance-level verification checklists. This shifts planning data construction from fixed benchmark collection to controllable generation while preserving realistic task grounding. We use PlanningBench to evaluate open-source and closed-source frontier LLMs, and find that current models still struggle to produce complete solutions under coupled constraints. Beyond evaluation, reinforcement learning on verified PlanningBench data improves performance on unseen planning benchmarks and broader instruction-following tasks. Further analysis suggests that determinate or well-specified optimal solutions provide clearer reward signals and more stable training dynamics. Overall, PlanningBench provides a controllable source of planning data for diagnosing and improving generalizable planning abilities in LLMs.\footnote{\url{https://github.com/Tencent-Hunyuan/PlanningBench}. Correspondence to zhaoziliang@ruc.edu.cn}
\end{abstract}

\begin{figure}[h]
    \centering
    \includegraphics[width=0.98\linewidth]{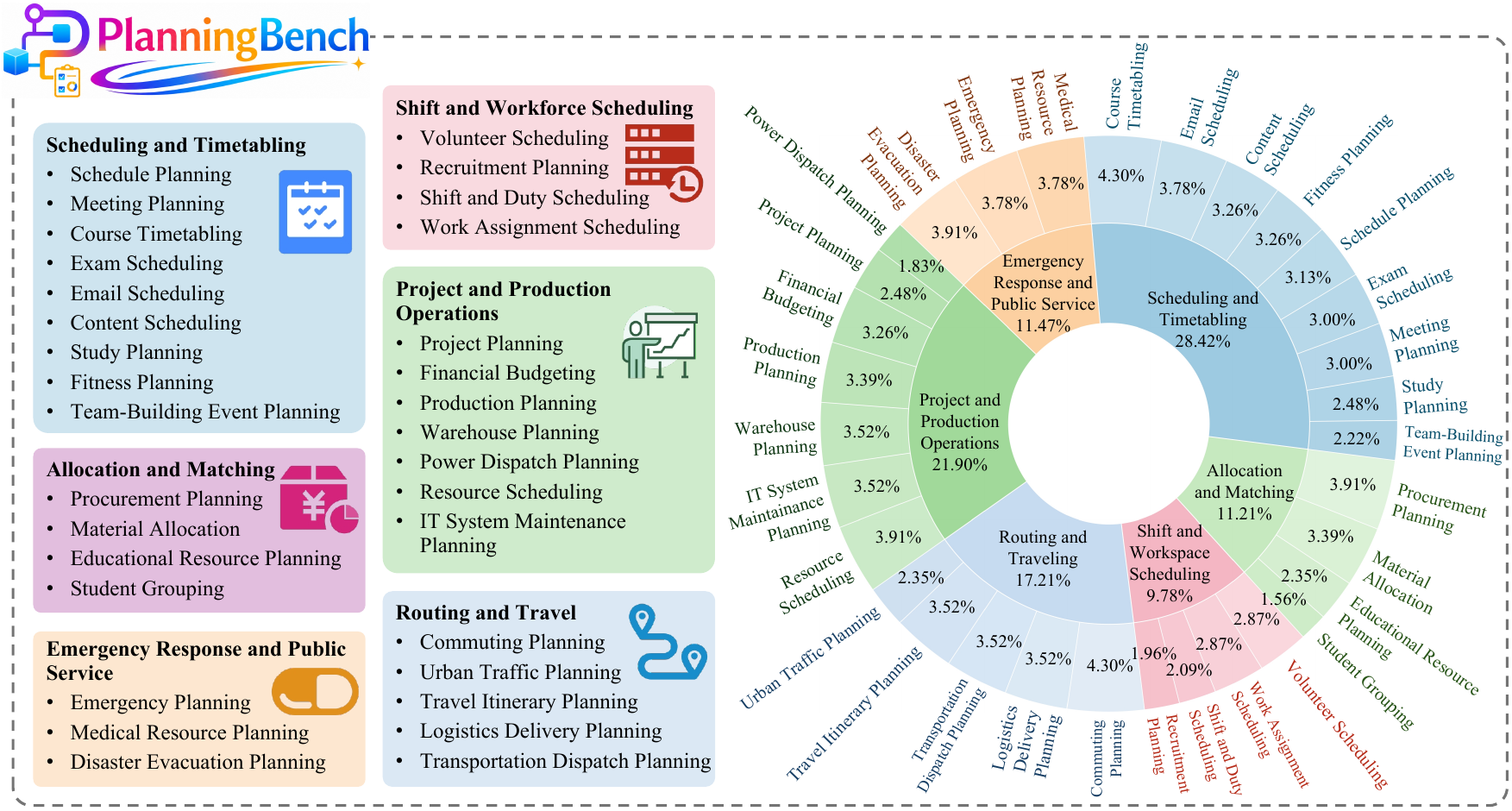}
    \caption{Overview of PlanningBench task taxonomy and data distribution, showing six high-level planning categories and the proportion of individual task types within each category.}
    \label{fig:figure1}
\end{figure}

\section{Introduction}
Planning has become a central challenge for evaluating and improving the ability of large language models (LLMs) to solve complex real-world problems. This capability reflects a foundational aspect of general intelligence because it requires models to reason about future consequences, allocate limited resources, and pursue long-term goals under constraints. Real-world planning tasks therefore provide a natural setting for both evaluation and training, as they require complete solutions that are executable and verifiable under coupled constraints~\citep{mitra2026recap, song2023llm, wei2025plangenllms}. These requirements distinguish planning from open-ended generation and localized reasoning benchmarks, where individual outputs or intermediate steps can often be evaluated in isolation~\citep{fan2024survey, setlur2024rl}. In planning, a locally reasonable decision may invalidate the global plan, and a feasible plan may still be far from optimal~\citep{dagan2023dynamic}. As frontier LLMs continue to improve, scalable sources of diverse and verifiable planning data are needed not only to faithfully evaluate their planning abilities, but also to further train models toward more generalizable planning behavior~\citep{deepplanning2026}.

A growing body of work has advanced LLM planning evaluation through realistic and verifiable benchmarks~\citep{tripbench2026,costbench2025,compass2025,travelbench2026,travelplanner2024,worldtravel2026,deepplanning2026}. Recent studies also show that planning data can support model development beyond static evaluation~\citep{triflow2025,vaiage2025,atlas2025,experienceynthesis2025}. These advances highlight the value of planning data, but also reveal limitations of benchmark-driven design. Existing efforts often define fixed collections of planning tasks rather than scalable mechanisms for generating data with high diversity, controlled difficulty, and automatic verification. This limits coverage and diagnostic value, since many benchmarks focus on a small number of domains and test only a limited range of planning structures~\citep{worldtravel2026,chinatravel2024,travelbench2026}. Difficulty is also often approximated by surface-level proxies such as task length or number of requirements, which may not capture structural sources of planning difficulty such as task-specific design, hierarchy, and constraint coupling. Consequently, existing benchmarks may not provide a fine-grained view of frontier LLM capability limits or reliable verification signals for planning-oriented training~\citep{travelplanner2024,huang2025reinforcement}.

To address these limitations, we propose \textbf{PlanningBench} as a synthetic planning data generation framework for evaluating and training LLM planning abilities. PlanningBench starts from real planning scenarios abstracted by domain experts. These abstractions organize practical planning workflows into a taxonomy of task types, subtasks, constraint families, and difficulty factors. The taxonomy covers more than 30 planning tasks, each with multiple subtasks and a rich pool of task-specific constraints that can be systematically composed into diverse problem instances. In this way, PlanningBench preserves realistic task grounding while shifting planning data construction from collecting fixed benchmark instances to controllable data generation.

PlanningBench implements this controllable generation process through a constraint-driven synthesis pipeline. The pipeline samples structured task-constraint configurations and instantiates each configuration as a self-contained planning problem. It then applies controlled difficulty enhancement and post-hoc quality filtering. These steps enable controlled diversity across domains, subtasks, constraints, and solution structures. They also support task-specific and hierarchical difficulty control by varying constraint tightness, resource scarcity, objective conflicts, subtask dependencies, and global coordination requirements rather than only surface complexity. Each instance is paired with automatic verification rules that check constraint satisfaction and objective quality. Figure~\ref{fig:main_figure} provides an overview of the PlanningBench construction pipeline.

With this generated and automatically verifiable planning data, we study LLM planning from both evaluation and training perspectives. We systematically test open-source and closed-source frontier LLMs to examine the current boundary of planning capability. Even the strongest evaluated models achieve limited solve rates on PlanningBench, showing that the generated evaluation suite remains far from saturated. We further analyze performance across task topics, prompt lengths, and the number of checklist items, revealing how scenario structure and constraint complexity affect planning success. For training, we use verified PlanningBench data for GRPO-based reinforcement learning and evaluate transfer beyond the training distribution. The trained models improve on unseen planning benchmarks such as TravelPlanner~\citep{travelplanner2024} and ChinaTravel~\citep{chinatravel2024}, with further gains on broader instruction-following and complex reasoning benchmarks, including Multi-Challenge~\citep{deshpande2025multichallenge}, Inverse IFEval~\citep{zhang2025inverse}, and Collie~\citep{yao2023collie}. These gains suggest that PlanningBench does not merely teach models to imitate fixed task templates. Instead, its verification-driven training signals encourage constraint integration, multi-step reasoning, and global consistency maintenance~\citep{pyatkin2025generalizing}. Our training analysis further suggests that planning instances with more determinate optimal solutions provide more stable and directional reward signals than instances that admit many locally acceptable answers.

Overall, this work makes three main contributions.

\begin{itemize}
    \item We introduce PlanningBench, a synthetic planning data generation framework grounded in real planning scenarios. It generates scalable, diverse, and verifiable planning data with broad task coverage, task-specific constraints, and controllable difficulty.

    \item We show that verified PlanningBench data can support planning-oriented reinforcement learning. Training on PlanningBench improves performance on unseen planning benchmarks and broader instruction-following and reasoning tasks. We further identify reward determinacy as an important factor for stable and transferable planning-oriented training.

    \item We systematically evaluate open-source and closed-source frontier LLMs on PlanningBench. The results show strong discriminative power across models while the generated evaluation suite remains challenging even for the strongest evaluated models.
\end{itemize}

\begin{table*}[t]
\centering
\footnotesize
\setlength{\tabcolsep}{5pt}
\renewcommand{\arraystretch}{1.15}
\caption{Feature-level comparison between PlanningBench and representative planning benchmarks.}
\label{tab:benchmark_comparison}
\begin{tabular*}{\linewidth}{@{\extracolsep{\fill}}lcccccc}
\toprule
\textbf{Benchmark} &
\textbf{\shortstack{Real-scenario\\derived}} &
\textbf{\shortstack{Domain\\Coverage}} &
\textbf{\shortstack{Tool-free\\Planning}} &
\textbf{\shortstack{Scalable\\difficulty control}} &
\textbf{\shortstack{Training\\validated}} &
\textbf{\shortstack{Determinate\\objectives}} \\
\midrule
TravelPlanner      & $\checkmark$ & 1 type  & $\times$ & $\times$ & $\times$ & $\times$ \\
TravelBench        & $\checkmark$ & 1 type  & $\times$ & $\times$ & $\times$ & $\times$ \\
TRIP-Bench         & $\checkmark$ & 1 type  & $\times$ & $\checkmark$ & $\checkmark$ & $\times$ \\
WorldTravel        & $\checkmark$ & 1 type  & $\times$ & $\times$ & $\times$ & $\times$ \\
DeepPlanning       & $\checkmark$ & 2 types & $\times$ & $\checkmark$ & $\times$ & $\checkmark$ \\
CostBench          & $\times$     & 1 type  & $\times$ & $\checkmark$ & $\times$ & $\times$ \\
COMPASS            & $\checkmark$ & 1 type  & $\times$ & $\checkmark$ & $\times$ & $\times$ \\
ReliabilityBench   & $\times$     & 4 types & $\times$ & $\checkmark$ & $\times$ & $\times$ \\
TripTailor         & $\checkmark$ & 1 type  & $\checkmark$ & $\times$ & $\times$ & $\times$ \\
ChinaTravel        & $\checkmark$ & 1 type  & $\times$ & $\times$ & $\times$ & $\times$ \\
TripCraft          & $\checkmark$ & 1 type  & $\checkmark$ & $\times$ & $\times$ & $\times$ \\
\midrule
\textbf{PlanningBench (ours)}
                   & $\checkmark$ & \textbf{30+ types}
                   & $\checkmark$ & $\checkmark$ & $\checkmark$ & $\checkmark$ \\
\bottomrule
\end{tabular*}
\end{table*}

\section{Related Work}

Recent work has introduced many benchmarks for evaluating LLM planning. A large body of work focuses on travel planning and its variants, including retrieval-based planning, realistic user requests, infeasible tasks, dynamic changes, reward construction, and multimodal or web-based environments~\citep{travelplanner2024,travelbench2026,tripbench2026,worldtravel2026,triptailor2025,chinatravel2024,tripcraft2025,tripscore2025,triptide2025,flextravelplanner2025}. Other benchmarks extend planning evaluation to verifiable constraints, cost-sensitive tool sequencing, progressive constraint revelation, retail and service scenarios, execution-based simulation, and agent reliability~\citep{deepplanning2026,costbench2025,compass2025,deng2025retail,yao2024tau,barres2025tau,qian2025userbench,valmeekam2023planbench,reliabilitybench2025,vitabench2025}. These benchmarks provide valuable testbeds, but they mostly treat planning data as fixed task collections or simulated interactions. Their difficulty is often controlled through general proxies such as prompt length, number of requirements, interaction depth, or tool-use complexity, which only indirectly capture structural sources of planning difficulty.

PlanningBench treats planning data construction as a controllable generation problem. It abstracts reusable task and constraint structures from real scenarios and synthesizes self-contained instances through task-specific constraint composition and hierarchical difficulty control. This design covers more than 30 task types and pairs each instance with verification checklists, allowing the same data to support complete-solution evaluation and planning-oriented training. In this sense, PlanningBench complements prior benchmarks by providing a scalable source of diverse and verifiable planning data rather than another fixed evaluation set. Table~\ref{tab:benchmark_comparison} compares representative benchmarks along these dimensions. Domain coverage counts task types or domains. Tool-free planning excludes external tools and environment interaction. Determinate objectives refer to unique, determinate, or well-specified optima.

Alongside benchmark development, recent methods improve LLM planning through task decomposition, multi-agent collaboration, constraint repair, reinforcement learning, human-in-the-loop workflows, and synthesized experience~\citep{triflow2025,vaiage2025,atlas2025,deeptravel2025,roamify2025,aligningagents2025,experienceynthesis2025,smartagents2024,travelplannerrely2024}. These methods mainly focus on designing stronger planners, agents, or tool-use systems. PlanningBench is complementary because it focuses on generating structured and verifiable planning data for evaluating and training models. Its Generator, Responder, and Critic are closed-loop synthesis components for data construction, not a new agent architecture.

\section{Data Construction}

\subsection{Definition and Scope of PlanningBench}

PlanningBench targets text-based planning problems in which all information needed to construct and verify a plan is provided in the input. Each generated instance is represented as \((x_i,c_i)\), where \(x_i\) denotes a self-contained planning task and \(c_i\) denotes the verification checklist used for evaluation and training~\citep{gunjal2025rubrics,liu2025openrubrics}. The checklist specifies the requirements a model-generated plan must satisfy, including relevant context, constraints, output conditions, and task objectives. Under this formulation, PlanningBench asks whether a model can integrate the provided information into a concrete, executable, and verifiable plan. When an instance specifies a determinate optimum, verification further checks whether the plan reaches that optimum.

\subsection{Overview of the Construction Pipeline}

PlanningBench constructs scalable and verifiable planning data through real-scenario abstraction, task-constraint taxonomy, and constraint-driven synthesis. It abstracts reusable planning schemas from representative planning scenarios, where each schema specifies the planning target, constraint space, and controllable difficulty factors.

\begin{figure}[tp]
    \centering
    \includegraphics[width=0.98\linewidth]{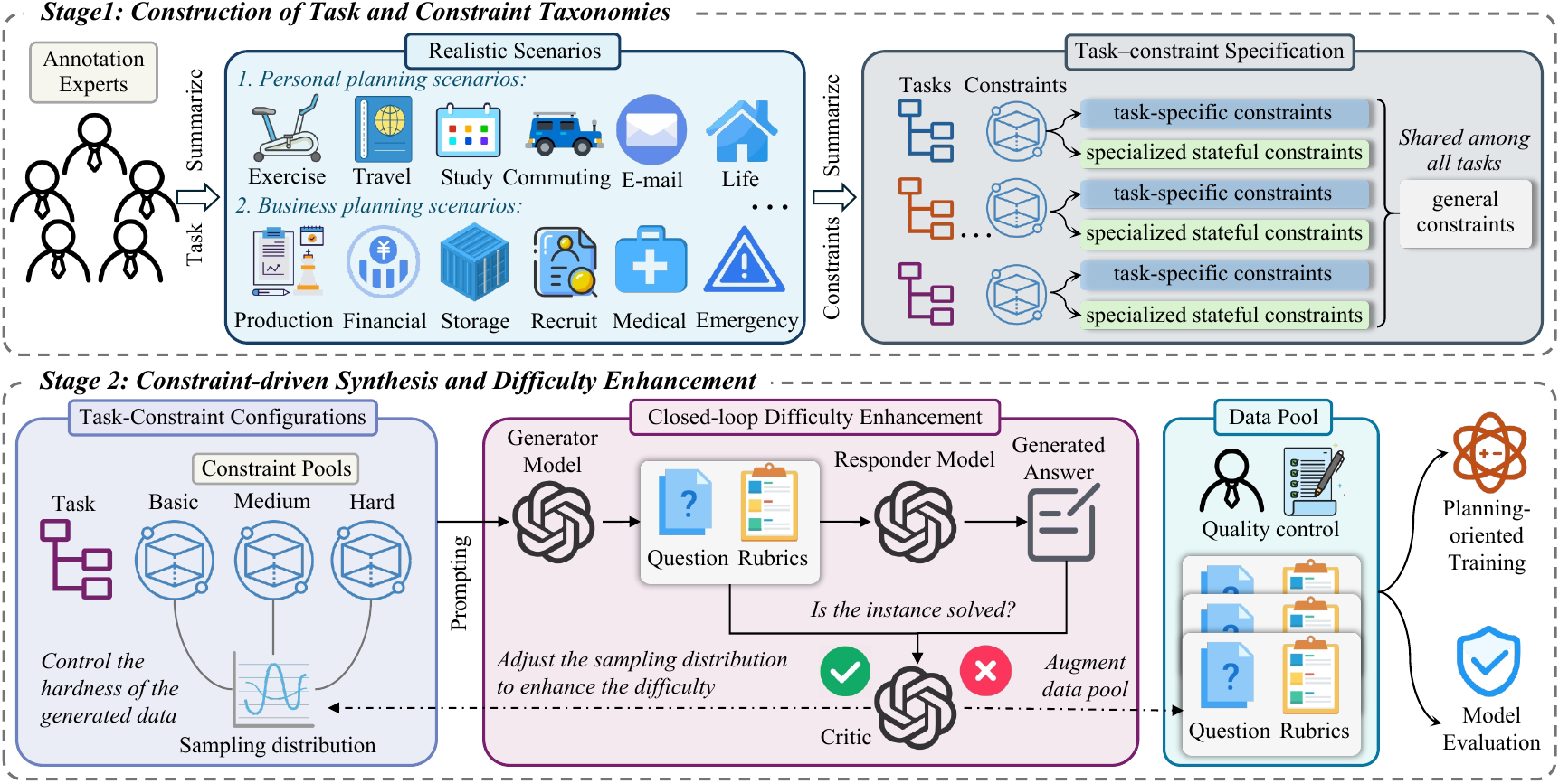}
    \caption{PlanningBench construction pipeline. PlanningBench first abstracts real planning scenarios into task and constraint taxonomies, then generates self-contained instances through constraint-driven closed-loop synthesis. Candidate instances are paired with checklists and filtered before being used for evaluation and training.}
    \label{fig:main_figure}
\end{figure}

As shown in Figure~\ref{fig:main_figure}, construction has two macro stages. The first builds task and constraint taxonomies from real scenarios, covering task types, subtask variants, general constraints, task-specific constraints, and specialized stateful constraints. The second synthesizes self-contained instances by sampling task-constraint configurations, attaching verification checklists, and adjusting difficulty through closed-loop feedback.

Candidate instances are further filtered and revised for clarity, consistency, and verifiability. The resulting verified data pool supports both model evaluation and planning-oriented training. This process preserves realistic task grounding while turning planning data construction from fixed collection into controllable generation.

\subsection{Task and Constraint Taxonomy}

The first macro stage converts representative planning scenarios into task and constraint taxonomies. Its purpose is not to catalog domains, but to define a reusable design space for planning data generation. To construct this space, twenty professional annotators with planning experience, together with algorithm researchers, review representative scenarios from real applications. They abstract stable problem structures, including planning targets, candidate actions, resources, time windows, dependencies, and required plan conditions. These structures are organized into two layers. The task taxonomy captures what type of plan to construct, while the constraint taxonomy captures feasibility, optimization, and state-dependent requirements that make the plan verifiable. Together, these taxonomies provide the structured elements from which PlanningBench samples and composes diverse planning instances.

\subsubsection{Task Taxonomy}

The task taxonomy is organized by planning structure rather than surface application labels. Based on annotated scenarios, we group planning tasks into six families that capture different sources of structural planning complexity:
\begin{enumerate}[label=(\arabic*),leftmargin=1.7em,itemsep=0.12em,topsep=0.3em]
    \item \emph{Scheduling and timetabling} covers temporal conflicts, time-window matching, and execution ordering.
    \item \emph{Allocation and matching} focuses on resource assignment under compatibility and capacity constraints.
    \item \emph{Shift and workforce scheduling} emphasizes coverage, rotation fairness, and workforce availability.
    \item \emph{Routing and travel} involves route selection, spatiotemporal coordination, and multi-leg transfers.
    \item \emph{Project and production operations} captures milestones, dependencies, capacity limits, and execution continuity.
    \item \emph{Emergency response and public service} focuses on timeliness, priority-based allocation, and plan restructuring.
\end{enumerate}
This organization allows PlanningBench to cover a systematic range of planning structures while keeping the generation space controllable. At a finer granularity, each family contains concrete tasks derived from planning scenarios, and each task is divided into 5--10 \emph{subtasks} on average to capture variants of the same planning problem type. A detailed task taxonomy with representative subtasks is provided in Appendix~\ref{app:task_taxonomy_summary} Table~\ref{tab:appendix_tasks}.

\subsubsection{Constraint Taxonomy}

Constraints define the feasibility and optimization requirements that make planning instances verifiable and controllable during synthesis. Because task and constraint structures are closely coupled, we model constraints together with task types rather than adding them as post-hoc annotations. This produces a structured task-constraint specification in which each task is associated with reusable constraint elements.

We organize the extracted constraints into three categories:
\begin{itemize}[leftmargin=1.2em,itemsep=0.12em,topsep=0.3em]
    \item \emph{General constraints} are shared across tasks and capture common planning requirements, such as complete input information, time-window satisfaction, resource boundaries, and infeasibility recognition.
    \item \emph{Task-specific constraints} are tied to a particular task structure and define requirements that arise from that task.
    \item \emph{Specialized stateful constraints} depend on previously executed actions, accumulated state, or triggered conditions.
\end{itemize}
At the taxonomy level, these are reusable \emph{meta-constraints} rather than rules tied to a single problem, allowing PlanningBench to scale generation beyond fixed manually written instances.

To support difficulty control, we organize constraints into three levels for each task. The \emph{basic} level covers fundamental feasibility requirements, including complete input, resource boundaries, time windows, capacity limits, and executability. The \emph{medium} level introduces optimization objectives, such as fairness, balance, multi-objective trade-offs, and load balancing. The \emph{hard} level captures demanding structural requirements, including infeasibility recognition, exception recovery, robustness design, and coordination under conflicting objectives. Specialized stateful constraints are maintained as a separate optional layer and sampled independently when needed.

During data construction, the pipeline samples a task and subtask variant, selects associated constraints, and instantiates the resulting configuration with difficulty control and closed-loop enhancement. In this way, the taxonomy connects real-scenario grounding with scalable synthetic data generation. Detailed constraint descriptions and representative examples are provided in Appendix~\ref{app:constraint_taxonomy_summary} Tables~\ref{tab:appendix_general_constraints} and~\ref{tab:appendix_task_specific_constraints}.

\subsection{Constraint-driven Synthesis and Difficulty Enhancement}
\label{subsec:constraint_synthesis}

Given the task and constraint taxonomies, PlanningBench synthesizes instances by instantiating structured task-constraint configurations as self-contained problems. We implement this process with three closed-loop components, namely a Generator, a Responder, and a Critic, as shown in Figure~\ref{fig:main_figure}. The Generator creates candidate instances, the Responder attempts solutions, and the Critic verifies them. This loop generates diverse and verifiable planning instances while increasing difficulty when the current Responder fully solves a generated problem.

For a sampled task \(\tau\) and subtask variant \(\sigma\), the Generator draws constraints from task-specific basic, medium, and hard constraint pools, denoted by \(\mathcal{C}_b(\tau)\), \(\mathcal{C}_m(\tau)\), and \(\mathcal{C}_h(\tau)\). To keep each instance manageable while allowing controlled variation in difficulty, the initial counts satisfy
\[
N_b\in\{1,2,3\}, \quad
N_m\in\{0,1,2\}, \quad
N_h\in\{0,1\}.
\]
After sampling the task, subtask, and constraints, the Generator instantiates a self-contained problem \(x\) and its associated verification checklist \(c\). Additional sampling and subset construction details are provided in Appendix~\ref{app:constraint_sampling}. Given the generated problem \(x\), the Responder outputs a candidate plan \(\hat{y}\). The Critic evaluates \(\hat{y}\) against the checklist and a prompt-based verifier, producing a verification score \(\rho\in[0,1]\) and a binary all-pass indicator
\[
u=\mathbb{I}\{\hat{y}\ \text{satisfies all constraints and passes final verification}\}.
\]
The indicator \(u\) is used as feedback for subsequent synthesis rather than as a final benchmark score. When \(u=1\), the current Responder fully solves the instance, so the pipeline keeps the same task and shifts the constraint composition toward higher difficulty. When \(u=0\), the instance remains challenging under the current configuration.

Let \(\mathbf{p}^{(k)}=(p_b^{(k)},p_m^{(k)},p_h^{(k)})\) denote the sampling probabilities over basic, medium, and hard constraints at iteration \(k\). When a harder variant is requested, the next distribution is updated by
\begin{equation}
\mathbf{p}^{(k+1)}=
\mathrm{Normalize}(
\mathbf{p}^{(k)} \odot \exp(\eta[-\alpha,\beta,\gamma])),
\qquad \alpha,\beta,\gamma,\eta>0 .
\end{equation}
This update decreases the weight of basic constraints and increases the weights of medium and hard constraints. The updated probabilities are then projected back to admissible discrete constraint counts before the next generation step, as detailed in Appendix~\ref{app:constraint_sampling}. Overall, PlanningBench performs adaptive search over structured planning configurations, turning the task and constraint taxonomies into a controllable mechanism for scalable planning data generation.

\subsection{Automatic Verification and Quality Control}
\label{subsec:verification_quality_control}

After the generation pipeline, each generated instance includes a verification checklist that converts prompt-level constraints and objectives into instance-level checks. These checks cover input conditions, resource and time constraints, output format, and determinate objectives when available. This design makes PlanningBench instances directly usable for both model evaluation and planning-oriented training, since the same verification structure can provide feedback and reward signals.

After automatic verification, we conduct a human quality-control audit to assess data usability. Twenty professional annotators review synthesized samples and assign each instance to one of four outcomes: direct retention, retention after minor revision, retention after source correction, or discard. Recoverable samples are revised and retained, while irrecoverable samples are removed. In the audited batch, 86.15\% of synthesized samples require no revision or only minor revision, and 13.85\% require additional source correction before retention. No sample in this batch is directly assigned to the discard category. These results suggest that the synthesis pipeline usually produces recoverable planning instances, while human revision further improves the clarity, consistency, and verifiability of the final data. Details of the audit categories and common revision types are provided in Appendix~\ref{app:quality_control}.

\subsection{Preference for Determinate Optimal Solutions}
\label{subsec:determinate_optima}

Beyond making each instance verifiable, PlanningBench favors determinate or well-specified optimal solutions whenever the task permits. This design principle is useful for both planning-oriented training and evaluation~\citep{su2025expanding,guo2025deepseek,xu2021fine}. In planning tasks, loose answer spaces can allow many outputs to be partially feasible. If verification rewards such outputs too generously, a model may learn local constraint satisfaction without learning to coordinate time, resources, dependencies, and objectives at the global level. The resulting reward signal can be positive but diffuse. It may indicate how to obtain partial credit, but not how to construct a globally consistent plan. A more concentrated answer space also strengthens evaluation by distinguishing models that merely satisfy local requirements from those that integrate constraints and optimize the full plan.

This preference is informed by early data construction and training practice. In an early batch of about 1k synthesized examples, GRPO training did not yield clear gains and degraded performance on general instruction-following benchmarks. We conjecture that loose optima and permissive verification made the reward less directional. Later batches therefore strengthen the preference for determinate or otherwise well-specified optimal solutions whenever possible. Subsequent evaluations show clearer gains, although more systematic experiments are still needed. We further analyze this effect in Section~\ref{subsec:planningbench_training}.

\section{Experiments}

We evaluate PlanningBench from two complementary perspectives. First, we study whether the generated data forms a challenging and discriminative evaluation suite for current LLMs. Second, we study whether the same automatically verifiable data can support planning-oriented reinforcement learning and transfer beyond the training distribution. These two perspectives correspond to the central goal of PlanningBench, which is to generate scalable, diverse, and verifiable planning data for both evaluation and training.

The experiments are organized as follows. Section~\ref{subsec:planningbench_evaluation} evaluates representative open-source and closed-source models on PlanningBench using All-pass and Avg-pass, and further analyzes performance variation and semantic failure patterns. Section~\ref{subsec:planningbench_training} studies whether PlanningBench data can support planning-oriented training by evaluating transfer to external planning benchmarks and broader instruction-following benchmarks.

\subsection{PlanningBench as an Evaluation Suite}
\label{subsec:planningbench_evaluation}

We first evaluate whether PlanningBench can serve as a challenging and diagnostic evaluation suite for structured planning. The goal is to test whether current LLMs can integrate all provided information into an executable and verifiable plan, rather than only satisfy isolated local requirements. Since planning failures often arise from global inconsistency rather than complete task misunderstanding, we evaluate both complete-solution success and partial checklist satisfaction. The following subsections report model performance, analyze factors that influence difficulty, and examine the semantic error types behind failed outputs.

\subsubsection{Evaluation Setup}

During synthetic evaluation data construction, we use strong models as the Generator to inform the human annotation process, through which we obtain the final evaluation set. We then us \texttt{Qwen-A3B-30B} as the Responder and \texttt{GPT-oss-120b} as the Critic model. We evaluate representative open-source and closed-source LLMs on the PlanningBench evaluation set, which contains 467 planning instances. All models follow the same evaluation protocol and are queried with default inference parameters unless otherwise specified. We use \texttt{GPT-oss-120b} as the judge model. For each instance, the judge assesses the model output against its verification checklist.

We report two complementary metrics. \textbf{All-pass} measures whether a plan satisfies all checklist items in a single response, capturing complete-solution success under coupled constraints. \textbf{Avg-pass} measures the fraction of checklist items satisfied by the output, capturing partial progress toward a valid plan. The gap is important because a model may satisfy many local requirements while still failing to produce a globally consistent solution.

\subsubsection{Performance of Open-Source and Closed-Source Models}

Table~\ref{tab:model_results_bar} reports the performance of representative LLMs on PlanningBench, with models ordered by All-pass. PlanningBench remains challenging even for the strongest evaluated models. \texttt{GPT-5.4-xhigh} achieves the best performance, with 63.17\% All-pass and 92.35\% Avg-pass, but still fails to satisfy the full verification checklist on more than one third of the evaluation set. \texttt{Gemini-3-1-pro} follows with 53.25\% All-pass and 88.36\% Avg-pass. Several other models, including \texttt{Seed-2.0-pro-high}, \texttt{DeepSeek-V3.2-thinking}, \texttt{Hy3-Preview}, and \texttt{Qwen3.5-plus-thinking}, achieve relatively high Avg-pass scores but substantially lower All-pass scores. These results indicate that the generated evaluation suite is far from saturated.

\begin{table*}[t]
\centering
\small
\caption{Performance of representative LLMs on PlanningBench. Models are ranked by All-pass, the primary metric. Higher is better. The best result is shown in bold and the second-best result is underlined.}
\label{tab:model_results_bar}
\setlength{\tabcolsep}{5pt}
\renewcommand{\arraystretch}{1.12}
\begin{tabularx}{\textwidth}{>{\raggedright\arraybackslash}p{3.6cm} >{\centering\arraybackslash}p{1.9cm} >{\centering\arraybackslash}p{0.9cm} >{\raggedright\arraybackslash}X >{\raggedright\arraybackslash}X}
\toprule
\textbf{Model} & \textbf{Type} & \textbf{Rank} & \textbf{All-pass (\%)} $\uparrow$ & \textbf{Avg-pass (\%)} $\uparrow$ \\
\midrule
GPT-5.4-xhigh             & Closed-source & 1  & \scorebar{63.17}{\best{63.17}}          & \scorebar{92.35}{\best{92.35}} \\
GPT-5.4-high              & Closed-source & 2  & \scorebar{58.56}{\second{58.56}}   & \scorebar{84.60}{84.60} \\
GPT-5.4-medium            & Closed-source & 3  & \scorebar{58.09}{58.09} & \scorebar{90.03}{\second{90.03}} \\
Gemini-3-1-pro            & Closed-source & 4  & \scorebar{53.25}{53.25}          & \scorebar{88.36}{88.36} \\
Seed-2.0-pro-high         & Open-source   & 5  & \scorebar{44.33}{44.33}          & \scorebar{84.02}{84.02} \\
DeepSeek-V3.2-thinking    & Open-source   & 6  & \scorebar{37.13}{37.13}          & \scorebar{80.21}{80.21} \\
Hy3-Preview               & Open-source   & 7  & \scorebar{36.17}{36.17}          & \scorebar{78.14}{78.14} \\
Qwen3.5-plus-thinking     & Open-source   & 8  & \scorebar{34.03}{34.03}          & \scorebar{77.10}{77.10} \\
DeepSeek-R1               & Open-source   & 9  & \scorebar{33.07}{33.07}          & \scorebar{74.04}{74.04} \\
Gemini-2.5-pro            & Closed-source & 10 & \scorebar{31.70}{31.70}          & \scorebar{76.41}{76.41} \\
Seed-2.0-pro-medium       & Open-source   & 11 & \scorebar{30.63}{30.63}          & \scorebar{77.09}{77.09} \\
DeepSeek-V3.2-exp         & Open-source   & 12 & \scorebar{29.27}{29.27}          & \scorebar{76.69}{76.69} \\
Hy2.1                     & Open-source   & 13 & \scorebar{24.61}{24.61}          & \scorebar{71.18}{71.18} \\
Seed-1.8                  & Open-source   & 14 & \scorebar{22.40}{22.40}          & \scorebar{71.69}{71.69} \\
Qwen3-30b-moe             & Open-source   & 15 & \scorebar{12.15}{12.15}          & \scorebar{57.40}{57.40} \\
Qwen3-32b                 & Open-source   & 16 & \scorebar{0.27}{0.27}            & \scorebar{30.11}{30.11} \\
Qwen3-14b                 & Open-source   & 17 & \scorebar{0.00}{0.00}            & \scorebar{25.69}{25.69} \\
Qwen3-8b                  & Open-source   & 18 & \scorebar{0.00}{0.00}            & \scorebar{22.79}{22.79} \\
\bottomrule
\end{tabularx}
\end{table*}

The gap between All-pass and Avg-pass shows that local constraint satisfaction does not necessarily imply complete planning success. For example, \texttt{GPT-5.4-medium} reaches 90.03\% Avg-pass but only 58.09\% All-pass. Similarly, \texttt{Seed-2.0-pro-high} reaches 84.02\% Avg-pass but only 44.33\% All-pass. These gaps suggest that models often satisfy many checklist items while still failing to produce a globally valid plan because of a small number of critical errors. PlanningBench therefore distinguishes partial compliance from the ability to integrate locally correct decisions into a complete and consistent plan.

For weaker models, exact solving ability drops sharply. \texttt{Qwen3-30b-moe} achieves only 12.15\% All-pass, while \texttt{Qwen3-32b}, \texttt{Qwen3-14b}, and \texttt{Qwen3-8b} are nearly unable to solve full instances under the All-pass metric. Their Avg-pass scores show that some local requirements can still be satisfied, but the strict global-solving metric exposes a larger capability gap. Overall, PlanningBench provides useful discrimination across model capability levels and reveals substantial remaining limitations in structured planning.

\subsubsection{Influencing Factor Analysis}

\begin{figure}[t]
    \centering
    \includegraphics[width=0.98\linewidth]{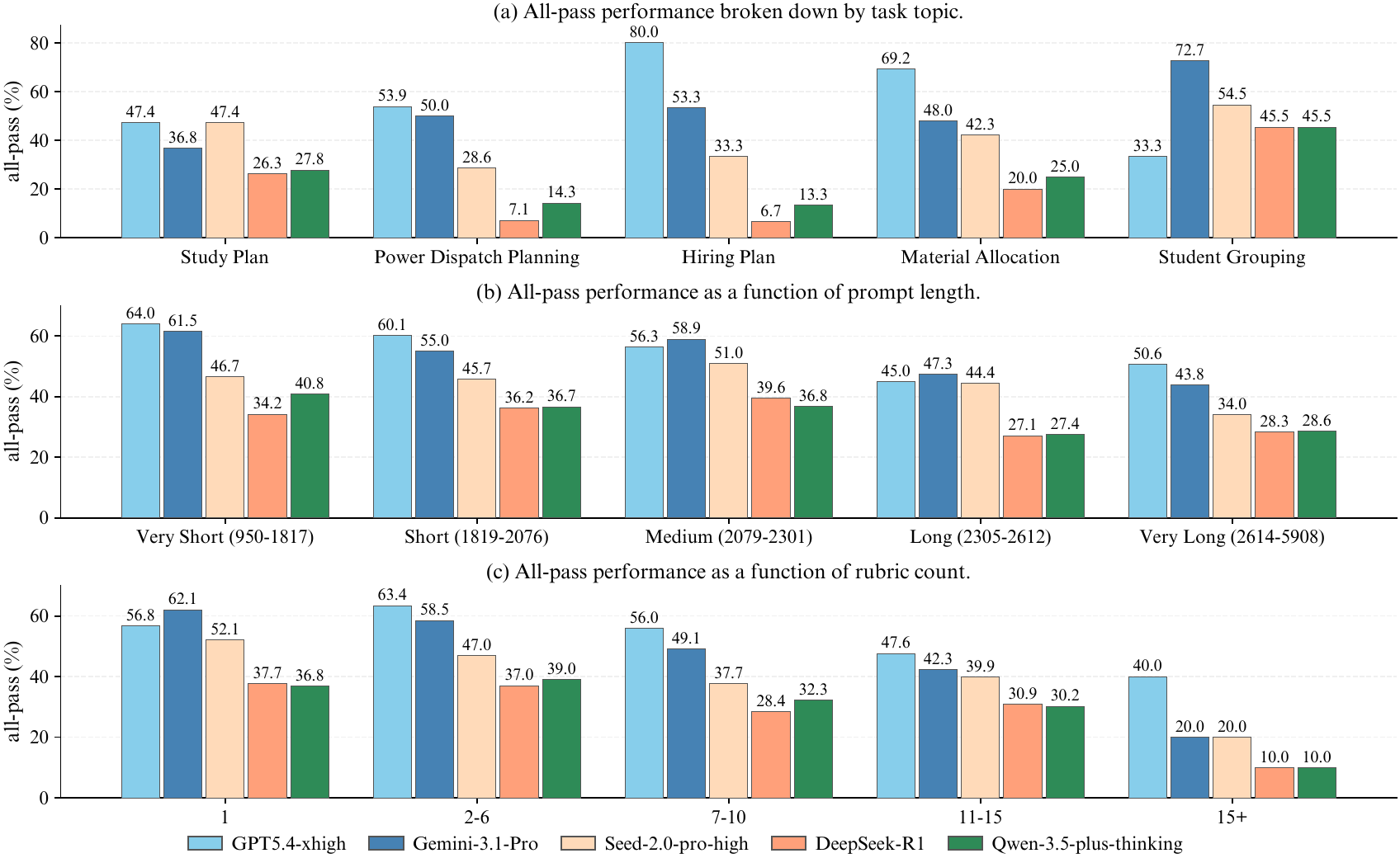}
    \caption{All-pass performance across task type, prompt length, and number of checklist items.}
    \label{fig:rubric_count}
\end{figure}

We further analyze model performance by task type, prompt length, and number of checklist items. These factors reflect planning structure, information load, and constraint density. They do not fully determine instance difficulty, but provide useful views of where PlanningBench becomes challenging. Figure~\ref{fig:rubric_count} shows the results.

Performance varies substantially across task types, and models show distinct task-specific profiles. For example, \texttt{GPT-5.4-xhigh} performs best on Hiring Plan and Material Allocation, reaching 80.0\% and 69.2\% All-pass. In contrast, \texttt{Gemini-3-1-pro} achieves its highest score on Student Grouping, with 72.7\% All-pass. Some task types also expose larger cross-model gaps. On Hiring Plan, performance ranges from 80.0\% to 6.7\%, while on Power Dispatch Planning, it ranges from 53.9\% to 7.1\%. These results suggest that PlanningBench captures task-dependent differences in models' ability to handle complex constraints and maintain planning consistency.

Longer prompts are generally more difficult, although prompt length is not explicitly controlled during construction. \texttt{GPT-5.4-xhigh} declines from 64.0\% in the Very Short group to 50.6\% in the Very Long group, and \texttt{Gemini-3-1-pro} drops from 61.5\% to 43.8\%. The trend is not strictly monotonic across intermediate buckets, but the overall pattern suggests that larger information load makes it harder to satisfy all requirements consistently.

A larger number of checklist items also makes All-pass harder to achieve, especially in the highest-count bucket. \texttt{GPT-5.4-xhigh} reaches 63.4\% All-pass in the 2--6 item group, but drops to 40.0\% in the 15+ item group. \texttt{Gemini-3-1-pro} similarly decreases from 62.1\% in the single-item group to 20.0\% in the 15+ item group. Other models show similar declines in this setting. Overall, these results suggest that PlanningBench difficulty reflects both information load and the need to jointly satisfy multiple constraints while preserving global coherence.

\subsubsection{Error Analysis}
\label{subsec:error_analysis}

To better understand model failures beyond aggregate pass rates, we analyze All-pass failure cases, excluding exact refusals and blank or no-answer outputs. Each remaining case is assigned to one primary semantic error type. Constraint Omitted refers to missing required constraints or details. Wrong Calculation / Assignment covers numerical, temporal, scheduling, allocation, or logical errors. State Tracking refers to failures to preserve or update an evolving solution. Format / Structure captures violations of the required output schema. Missing Rationale refers to missing explanations, comparisons, or verification steps when required. The distribution is reported in Table~\ref{tab:error_analysis}, with a representative failure case shown in Appendix~\ref{app:break_case} Table~\ref{tab:break_case_production_planning}.

\begin{table*}[t]
\centering
\footnotesize
\caption{Distribution of semantic error types across representative models on PlanningBench. Percentages are computed over semantic failures only, after excluding all-pass cases and blank/no-answer outputs. Each failed case is assigned to one primary error type.}
\label{tab:error_analysis}
\setlength{\tabcolsep}{3pt}
\renewcommand{\arraystretch}{1.08}
\begin{tabular}{@{}p{3.15cm}p{2.2cm}p{2.2cm}p{2.2cm}p{2.2cm}p{2.2cm}@{}}
\toprule
\multicolumn{1}{c}{\textbf{Model}}
& \multicolumn{1}{c}{\shortstack{\textbf{Constraint}\\\textbf{Omitted (\%)}}}
& \multicolumn{1}{c}{\shortstack{\textbf{Wrong Calc. /}\\\textbf{Assign. (\%)}}}
& \multicolumn{1}{c}{\shortstack{\textbf{State}\\\textbf{Tracking (\%)}}}
& \multicolumn{1}{c}{\shortstack{\textbf{Format /}\\\textbf{Structure (\%)}}}
& \multicolumn{1}{c}{\shortstack{\textbf{Missing}\\\textbf{Rationale (\%)}}} \\
\midrule
GPT-5.4-xhigh         & \scorebarerr{15.1}{15.1} & \scorebarerr{65.5}{65.5} & \scorebarerr{11.3}{11.3} & \scorebarerr{3.4}{3.4} & \scorebarerr{4.6}{4.6} \\
Gemini-3-1-pro        & \scorebarerr{19.9}{19.9} & \scorebarerr{60.9}{60.9} & \scorebarerr{8.7}{8.7}  & \scorebarerr{2.7}{2.7} & \scorebarerr{7.7}{7.7} \\
Seed-2.0-pro-high     & \scorebarerr{14.8}{14.8} & \scorebarerr{67.1}{67.1} & \scorebarerr{6.3}{6.3}  & \scorebarerr{1.2}{1.2} & \scorebarerr{10.7}{10.7} \\
Hy3-Preview           & \scorebarerr{10.7}{10.7} & \scorebarerr{74.3}{74.3} & \scorebarerr{7.5}{7.5}  & \scorebarerr{1.2}{1.2} & \scorebarerr{6.3}{6.3} \\
Qwen3.5-plus-thinking & \scorebarerr{14.2}{14.2} & \scorebarerr{70.1}{70.1} & \scorebarerr{5.3}{5.3}  & \scorebarerr{2.6}{2.6} & \scorebarerr{7.9}{7.9} \\
DeepSeek-R1           & \scorebarerr{11.6}{11.6} & \scorebarerr{73.2}{73.2} & \scorebarerr{8.3}{8.3}  & \scorebarerr{1.9}{1.9} & \scorebarerr{5.0}{5.0} \\
Gemini-2.5-pro        & \scorebarerr{11.7}{11.7} & \scorebarerr{71.0}{71.0} & \scorebarerr{8.5}{8.5}  & \scorebarerr{2.7}{2.7} & \scorebarerr{6.1}{6.1} \\
Hy2.1                 & \scorebarerr{11.4}{11.4} & \scorebarerr{74.0}{74.0} & \scorebarerr{7.1}{7.1}  & \scorebarerr{0.9}{0.9} & \scorebarerr{6.7}{6.7} \\
Seed-1.8              & \scorebarerr{9.4}{9.4}   & \scorebarerr{75.0}{75.0} & \scorebarerr{7.2}{7.2}  & \scorebarerr{1.3}{1.3} & \scorebarerr{7.0}{7.0} \\
Qwen3-30b-moe         & \scorebarerr{12.0}{12.0} & \scorebarerr{75.7}{75.7} & \scorebarerr{6.4}{6.4}  & \scorebarerr{0.7}{0.7} & \scorebarerr{5.2}{5.2} \\
Qwen3-32b             & \scorebarerr{4.2}{4.2}   & \scorebarerr{83.5}{83.5} & \scorebarerr{6.3}{6.3}  & \scorebarerr{0.9}{0.9} & \scorebarerr{5.1}{5.1} \\
\bottomrule
\end{tabular}
\end{table*}

Wrong Calculation / Assignment dominates across all models, accounting for 60.9\%--83.5\% of semantic errors. This indicates that the main bottleneck is often not format following, but making correct numerical, temporal, scheduling, allocation, and logical decisions under constraints. Constraint Omitted accounts for 4.2\%--19.9\% of failures, State Tracking for 5.3\%--11.3\%, Format / Structure for only 0.7\%--3.4\%, and Missing Rationale for 4.6\%--10.7\%.

Model-specific patterns further clarify these failures. \texttt{Gemini-3-1-pro} has the highest share of Constraint Omitted errors at 19.9\%, while \texttt{GPT-5.4-xhigh} has the highest State Tracking rate at 11.3\%. \texttt{Qwen3-32b} concentrates most failures in Wrong Calculation / Assignment at 83.5\%. Overall, PlanningBench exposes weaknesses in constrained reasoning, multi-constraint integration, and stateful plan revision, rather than formatting compliance.

\subsection{PlanningBench as Training Data}
\label{subsec:planningbench_training}

Beyond evaluation, PlanningBench is designed to provide automatically verifiable planning data for training. We study whether GRPO-based reinforcement learning on verified PlanningBench data improves performance beyond the training distribution. This section evaluates transfer to external planning benchmarks and broader instruction-following benchmarks, and then analyzes training dynamics under different data construction settings. The goal is not to show that planning data improves all general abilities uniformly, but to test whether verifiable planning data provides useful reward signals for constraint integration, multi-step reasoning, and global consistency.

\subsubsection{Training Setup}

We use \texttt{Qwen-A3B-30B} as the base model and conduct GRPO-based reinforcement learning~\citep{shao2024deepseekmath} on 300 PlanningBench training instances. We compare four settings. \texttt{Base Model} denotes the original model without reinforcement learning. \texttt{Syn-PlanningBench} uses PlanningBench data constructed with verification-oriented constraints and a preference for determinate or well-specified optimal solutions. \texttt{Syn-NotDetOptimal} uses synthetic planning data that does not emphasize determinate optimality to the same extent. To contextualize the effect of structured synthesis, we additionally include \texttt{Human-Authored} as a comparison baseline. This baseline is independently written from scratch by the same pool of twenty professional annotators, without using the PlanningBench taxonomy, constraint-driven synthesis pipeline, or generated instances. The three reinforcement learning datasets are controlled to be of comparable size.

We evaluate trained models on external planning and general instruction-following benchmarks. The planning benchmarks include ChinaTravel~\citep{chinatravel2024} and TravelPlanner~\citep{travelplanner2024}. ChinaTravel contains Easy, Medium, and Human subsets, and we refer to the last as ChinaTravel-Human to avoid confusion with \texttt{Human-Authored}. TravelPlanner contains TP-Train, TP-Val, and TP-Test. The general benchmarks include Multi-Challenge~\citep{deshpande2025multichallenge}, Inverse IFEval~\citep{zhang2025inverse}, and Collie~\citep{yao2023collie}, which test instruction integration, constraint following, and global consistency. Detailed training hyperparameters are reported in Appendix~\ref{app:training_details}.

\subsubsection{Generalization to External Planning Benchmarks}

We first evaluate planning-specific transfer on ChinaTravel~\citep{chinatravel2024} and TravelPlanner~\citep{travelplanner2024}. Table~\ref{tab:generalization_planning} reports Avg-pass and All-pass on both benchmarks. This comparison tests whether reinforcement learning on PlanningBench data improves planning performance beyond the training distribution.

\begin{table*}[t]
\centering
\footnotesize
\setlength{\tabcolsep}{2.8pt}
\caption{Generalization on two external planning benchmarks (\%). $^\dagger$ indicates statistically significant improvements over the base model with $p < 0.05$.}
\label{tab:generalization_planning}

\resizebox{\textwidth}{!}{%
\begin{tabular}{@{}llccc@{\hspace{10pt}}lccc@{}}
\toprule
& \multicolumn{4}{c}{\textbf{ChinaTravel} $\uparrow$} & \multicolumn{4}{c}{\textbf{TravelPlanner} $\uparrow$} \\
\cmidrule(lr){2-5} \cmidrule(lr){6-9}
\textbf{Metric} & \textbf{Subset} & \textbf{Base} & \textbf{Human-Authored} & \textbf{Syn-PlanningBench}
& \textbf{Subset} & \textbf{Base} & \textbf{Human-Authored} & \textbf{Syn-PlanningBench} \\
\midrule
Avg-pass & Easy & 71.77 & \deltainline{\underline{73.40}$^\dagger$}{+1.63} & \deltainline{\textbf{76.77}$^\dagger$}{+5.00}
& TP-Train & 81.52 & \deltainline{\underline{89.87}$^\dagger$}{+8.35} & \deltainline{\textbf{93.00}$^\dagger$}{+11.48} \\
All-pass & Easy & 14.63 & \deltainline{\underline{17.81}$^\dagger$}{+3.18} & \deltainline{\textbf{29.12}$^\dagger$}{+14.49}
& TP-Train & 29.80 & \deltainline{\underline{35.56}$^\dagger$}{+5.76} & \deltainline{\textbf{48.84}$^\dagger$}{+19.04} \\
\midrule
Avg-pass & Medium & 87.96 & \deltainline{\underline{87.96}}{+0.00} & \deltainline{\textbf{90.50}$^\dagger$}{+2.54}
& TP-Val & 76.89 & \deltainline{\underline{88.53}$^\dagger$}{+11.64} & \deltainline{\textbf{91.04}$^\dagger$}{+14.15} \\
All-pass & Medium & 51.37 & \deltainline{\underline{51.02}}{-0.35} & \deltainline{\textbf{52.78}$^\dagger$}{+1.41}
& TP-Val & 21.04 & \deltainline{\underline{31.11}$^\dagger$}{+10.07} & \deltainline{\textbf{43.82}$^\dagger$}{+22.78} \\
\midrule
Avg-pass & Human & 94.92 & \deltainline{\underline{95.59}}{+0.67} & \deltainline{\textbf{97.10}}{+2.18}
& TP-Test & 84.11 & \deltainline{\underline{88.56}$^\dagger$}{+4.45} & \deltainline{\textbf{91.76}$^\dagger$}{+7.65} \\
All-pass & Human & 86.76 & \deltainline{\underline{88.41}}{+1.65} & \deltainline{\textbf{93.18}$^\dagger$}{+6.42}
& TP-Test & \underline{35.72} & \deltainline{34.91}{-0.81} & \deltainline{\textbf{47.93}$^\dagger$}{+12.21} \\
\midrule
Avg-pass & Avg. & 84.88 & \deltainline{\underline{85.65}}{+0.77} & \deltainline{\textbf{88.12}$^\dagger$}{+3.24}
& Avg. & 80.84 & \deltainline{\underline{88.99}$^\dagger$}{+8.15} & \deltainline{\textbf{91.93}$^\dagger$}{+11.09} \\
All-pass & Avg. & 50.92 & \deltainline{\underline{52.41}$^\dagger$}{+1.49} & \deltainline{\textbf{58.36}$^\dagger$}{+7.44}
& Avg. & 28.85 & \deltainline{\underline{33.86}$^\dagger$}{+5.01} & \deltainline{\textbf{46.86}$^\dagger$}{+18.01} \\
\bottomrule
\end{tabular}
}
\end{table*}

\texttt{Syn-PlanningBench} consistently improves over \texttt{Base Model} on both external planning benchmarks. On ChinaTravel, Avg-pass increases from 84.88\% to 88.12\%, and All-pass from 50.92\% to 58.36\%. Even on the ChinaTravel-Human subset, where the base model performs strongly, \texttt{Syn-PlanningBench} improves All-pass from 86.76\% to 93.18\%. The transfer effect is stronger on TravelPlanner, where Avg-pass increases from 80.84\% to 91.93\% and All-pass from 28.85\% to 46.86\%. All-pass gains are consistent across TP-Train, TP-Val, and TP-Test, with the largest on TP-Val at +22.78 points. These results show that PlanningBench training transfers to external planning benchmarks, especially when tasks require stronger constraint coupling and global consistency.

Compared with \texttt{Human-Authored}, \texttt{Syn-PlanningBench} shows stronger transfer on both benchmarks. On ChinaTravel, \texttt{Human-Authored} brings 1.49 points of average All-pass improvement, compared with 7.44 points from \texttt{Syn-PlanningBench}. On TravelPlanner, \texttt{Human-Authored} reaches 33.86\% average All-pass, while \texttt{Syn-PlanningBench} reaches 46.86\%. Although \texttt{Human-Authored} provides valid training data, its examples are not constructed through taxonomy-guided constraint composition or controlled difficulty progression. This comparison highlights the value of structured and verification-oriented synthesis for planning-oriented training.

\subsubsection{Transfer to General Instruction-following Benchmarks}

We further evaluate whether PlanningBench training transfers beyond planning-specific benchmarks. Table~\ref{tab:generalization_general} reports results on Multi-Challenge, Inverse IFEval, and Collie. These benchmarks are not planning benchmarks in the same form as ChinaTravel or TravelPlanner, but they require models to integrate instructions, track constraints, and maintain consistency across complex responses.

\begin{table}[t]
\centering
\footnotesize
\setlength{\tabcolsep}{2.8pt}
\caption{Generalization on general-purpose instruction-following benchmarks (\%). $^\dagger$ indicates statistically significant improvements over the base model with $p < 0.05$.}
\label{tab:generalization_general}
\begin{tabularx}{0.95\linewidth}{
l
>{\centering\arraybackslash}X
>{\centering\arraybackslash}X
>{\centering\arraybackslash}X
>{\centering\arraybackslash}X}
\toprule
\textbf{Training Data} & \textbf{Multi-Challenge} $\uparrow$ & \textbf{Inverse IFEval} $\uparrow$ & \textbf{Collie} $\uparrow$ & \textbf{Avg.} $\uparrow$ \\
\midrule
\texttt{Base Model} & 29.18 & 48.72 & 38.33 & 38.74 \\
\texttt{Human-Authored} 
& \deltainline{\textbf{33.09}$^\dagger$}{+3.91}
& \deltainline{\underline{50.00}}{+1.28}
& \deltainline{\underline{42.33}$^\dagger$}{+4.00}
& \deltainline{\underline{41.81}$^\dagger$}{+3.07} \\
\texttt{Syn-NotDetOptimal} 
& \deltainline{\underline{30.28}}{+1.10}
& \deltainline{48.02}{-0.70}
& \deltainline{40.17}{+1.84}
& \deltainline{39.49}{+0.75} \\
\texttt{Syn-PlanningBench}
& \deltainline{\textbf{33.09}$^\dagger$}{+3.91}
& \deltainline{\textbf{51.14}$^\dagger$}{+2.42}
& \deltainline{\textbf{53.17}$^\dagger$}{+14.84}
& \deltainline{\textbf{45.80}$^\dagger$}{+7.06} \\
\bottomrule
\end{tabularx}
\end{table}

Overall, \texttt{Syn-PlanningBench} achieves the strongest transfer performance. It obtains the best or tied-best scores on all three benchmarks, improving the average score from 38.74\% to 45.80\%, with a gain of 7.06 points over \texttt{Base Model}. The improvement is especially pronounced on Collie, where the score increases from 38.33\% to 53.17\%. These results suggest that verified PlanningBench data can provide training signals that transfer beyond planning-specific evaluation settings.

The contrast with \texttt{Syn-NotDetOptimal} further suggests that broader transfer depends on the quality of the planning signal. \texttt{Syn-NotDetOptimal} brings only a marginal average gain of 0.75 points and slightly decreases performance on Inverse IFEval. This indicates that synthetic planning data without sufficiently clear optimality and verification signals provides less stable transfer. These results do not imply that planning data improves all general capabilities uniformly. Rather, they suggest that verification-driven PlanningBench data can provide useful training signals for tasks where models must coordinate multiple requirements across a complex response.

\subsubsection{Training Dynamics and the Role of Determinate Optima}

\begin{figure}[t]
    \centering
    \includegraphics[width=0.98\linewidth]{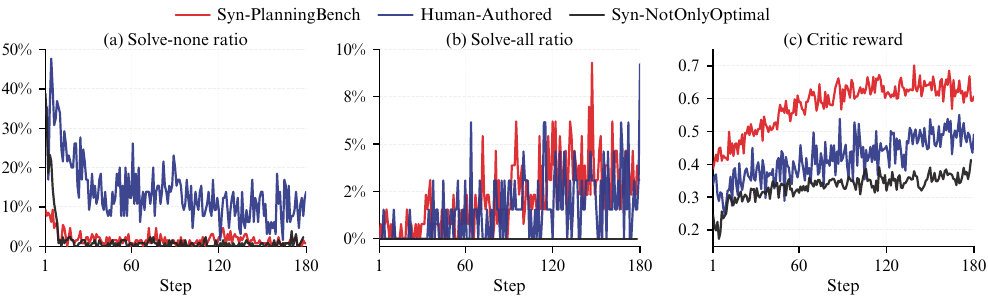}
    \caption{Training dynamics under three data types, \texttt{Human-Authored}, \texttt{Syn-NotDetOptimal}, and \texttt{Syn-PlanningBench}, measured by solve-none ratio, solve-all ratio, and critic reward.}
    \label{fig:training_dynamics}
\end{figure}

To examine how different data construction strategies affect reinforcement learning effectiveness, we compare the training dynamics of three models trained with different data sources mentioned above: \texttt{Human-Authored}, \texttt{Syn-NotDetOptimal}, and \texttt{Syn-PlanningBench}, in terms of solve-none ratio, solve-all ratio, and critic reward, as shown in Figure~\ref{fig:training_dynamics}. \texttt{Syn-PlanningBench} shows the most favorable dynamics. Its solve-none ratio decreases fastest and converges to the lowest level, while its solve-all ratio increases more consistently. Its critic reward curve is also smoother, suggesting more stable optimization signals. In contrast, \texttt{Syn-NotDetOptimal} shows weaker dynamics and maintains a low solve-all ratio, suggesting that partial checklist satisfaction does not reliably translate into complete solutions. \texttt{Human-Authored} performs between the two synthetic settings.

These trends align with the design preference in Section~\ref{subsec:determinate_optima}. Planning tasks require global consistency across many decisions, so permissive rewards may improve partial checklist satisfaction without reliably producing complete solutions. By favoring determinate or well-specified optima, \texttt{Syn-PlanningBench} provides clearer reward direction for complete planning success. Together with the transfer results, these dynamics suggest that reward determinacy is important for stable and transferable planning-oriented reinforcement learning.

\section{Conclusion}

We introduced PlanningBench as a synthetic planning data generation framework for LLM planning. Starting from real planning scenarios, it abstracts reusable task and constraint structures for constraint-driven synthesis, adaptive difficulty control, automatic verification, and quality filtering. This process generates scalable, diverse, and verifiable planning data for both evaluation and planning-oriented training. Experiments show that PlanningBench forms a challenging and discriminative evaluation suite, with even the strongest models leaving a substantial gap under the All-pass metric. Error analysis suggests that failures mainly reflect constrained reasoning and global consistency limitations rather than formatting issues. Beyond evaluation, GRPO-based training on verified PlanningBench data improves transfer to external planning and broader instruction-following benchmarks. The results further suggest that determinate or well-specified optimal solutions provide more stable and directional reward signals. Taken together, these findings show that controllable planning data construction, automatic verifiability, and reward determinacy are important for diagnosing and improving LLM planning abilities.

\bibliographystyle{iclr2025_conference}
\bibliography{iclr2025_conference}

\newpage

\appendix

\section{Task Taxonomy Summary}
\label{app:task_taxonomy_summary}

This section summarizes the task taxonomy underlying PlanningBench. Table~\ref{tab:appendix_tasks} lists the main task families, representative subtasks, and the primary abilities each task is intended to test, providing a concise supplement to the task-construction discussion in the main text.

{\scriptsize
\setlength{\tabcolsep}{2.2pt}
\renewcommand{\arraystretch}{1.06}
\setlength{\LTleft}{0pt}
\setlength{\LTright}{0pt}
\setlength{\emergencystretch}{4em}
\begin{longtable}{@{}>{\RaggedRight\hspace{0pt}\arraybackslash}p{0.16\linewidth}>{\RaggedRight\hspace{0pt}\arraybackslash}p{0.18\linewidth}>{\RaggedRight\hspace{0pt}\arraybackslash}p{0.30\linewidth}>{\RaggedRight\hspace{0pt}\arraybackslash}p{0.30\linewidth}@{}}
\caption{PlanningBench task taxonomy.}\label{tab:appendix_tasks}\\
\toprule
\textbf{Family} & \textbf{Main task} & \textbf{Illustrative subtask variants} & \textbf{What the task mainly tests} \\
\midrule
\endfirsthead

\toprule
\textbf{Family} & \textbf{Main task} & \textbf{Illustrative subtask variants} & \textbf{What the task mainly tests} \\
\midrule
\endhead

\midrule
\multicolumn{4}{r}{\scriptsize Continued on next page}\\
\endfoot

\bottomrule
\endlastfoot
Scheduling and Timetabling & Schedule Planning & one-day personal schedule; reprioritized agenda under urgent tasks & Arrange personal or team schedules under time windows and priorities. \\
Scheduling and Timetabling & Meeting Planning & cross-team meeting coordination; interview panel scheduling & Find shared meeting slots while handling room constraints, priorities, and rescheduling. \\
Scheduling and Timetabling & Course Timetabling & single-class weekly timetable; multi-class conflict-free timetabling & Resolve teacher/classroom/class conflicts under availability, capacity, and curriculum preferences. \\
Scheduling and Timetabling & Exam Scheduling & multi-course exam timetable; conflict-free room assignment across classes & Coordinate exam times, rooms, invigilators, and student load without conflicts. \\
Scheduling and Timetabling & Email Scheduling & customer notification schedule; recurring internal newsletter plan & Schedule outbound emails under audience timing, frequency, and campaign ordering constraints. \\
Scheduling and Timetabling & Content Scheduling & weekly social-media calendar; monthly short-video release plan & Plan multi-step content release pipelines under platform cadence and production dependencies. \\
Scheduling and Timetabling & Study Planning & exam-cram study plan; multi-subject final-week review schedule & Allocate study time across subjects under daily limits, priorities, and fixed commitments. \\
Scheduling and Timetabling & Fitness Planning & weekly fat-loss program; muscle-gain split routine & Design training schedules under goals, time limits, and physical restrictions. \\
Scheduling and Timetabling & Team-Building Event Planning & half-day team-building agenda; annual event program schedule & Arrange event programs, venues, and logistics under time, budget, and participant changes. \\
\midrule
Project and Production Operations & Project Planning & new-product milestone plan; budget-constrained project roadmap & Plan milestone-driven projects under staffing, budget, dependencies, and delivery deadlines. \\
Project and Production Operations & Financial Budgeting & departmental quarterly budget; event-level budget split & Distribute budgets across competing needs under caps, priorities, and reserve requirements. \\
Project and Production Operations & Production Planning & single-line production schedule; priority ordering across multiple orders & Schedule orders on constrained production lines under materials, changeovers, and urgency. \\
Project and Production Operations & Warehouse Planning & warehouse storage layout; inbound sequencing under capacity limits & Assign storage locations under zoning, capacity, accessibility, and turnover constraints. \\
Project and Production Operations & Power Dispatch Planning & daily city power plan; dispatch during peak demand & Allocate power supply under demand peaks, generation limits, and critical-load protection. \\
Project and Production Operations & Resource Scheduling & multi-task resource schedule; sequencing for shared equipment & Schedule shared people/equipment across concurrent tasks under cost or makespan objectives. \\
Project and Production Operations & IT System Maintenance Planning & system-upgrade maintenance window; low-traffic maintenance plan & Sequence maintenance windows and rollback-ready changes under service and dependency constraints. \\
\midrule
Allocation and Matching & Procurement Planning & monthly office procurement plan; restocking under low inventory & Choose replenishment timing and suppliers under lead time, inventory, and budget constraints. \\
Allocation and Matching & Material Allocation & limited office-supply allocation; priority-based relief allocation & Allocate scarce supplies across units under fairness, priority, and minimum-guarantee rules. \\
Allocation and Matching & Educational Resource Planning & teacher/classroom assignment; shared teaching-resource planning & Assign teachers, classrooms, and equipment across classes under shared-resource constraints. \\
Allocation and Matching & Student Grouping & balanced student grouping; mixed-ability grouping & Form balanced groups under size, compatibility, and diversity requirements. \\
\midrule
Emergency Response and Public Service & Emergency Planning & fire-response workflow; extreme-weather emergency plan & Plan emergency response flows under deadlines, roles, materials, and concurrent incidents. \\
Emergency Response and Public Service & Medical Resource Planning & outpatient appointment plan; doctor duty and clinic scheduling & Schedule doctors, rooms, and patient flow under urgency, capacity, and waiting-time targets. \\
Emergency Response and Public Service & Disaster Evacuation Planning & building evacuation routing; school emergency evacuation plan & Design safe evacuation plans under exit capacity, vulnerable groups, and route failures. \\
\midrule
Shift and Workforce Scheduling & Volunteer Scheduling & event volunteer post assignment; staffing under limited volunteers & Deploy volunteers to posts under skills, training, peak demand, and rotation constraints. \\
Shift and Workforce Scheduling & Recruitment Planning & multi-role hiring pipeline; multi-round interview schedule & Plan multi-stage hiring pipelines under interviewer availability, urgency, and budget limits. \\
Shift and Workforce Scheduling & Shift and Duty Scheduling & hospital nurse roster; customer-service duty schedule & Build fair rosters with coverage, rest, qualification, and holiday rules. \\
Shift and Workforce Scheduling & Work Assignment Scheduling & weekly team task allocation; deadline-driven reprioritization & Assign tasks to staff under skills, deadlines, dependencies, and balance. \\
\midrule
Routing and Travel & Commuting Planning & backup commute in bad weather; commute plus child drop-off & Plan commute routes under arrival-time, cost, transfer, and contingency constraints. \\
Routing and Travel & Urban Traffic Planning & peak-hour traffic diversion; traffic plan for large events & Coordinate routing/diversion schemes under road capacity, peaks, and event-driven demand. \\
Routing and Travel & Travel Itinerary Planning & 3-day short trip; low-intensity family itinerary & Construct itineraries under budget, reservations, preferences, mobility limits, and disruptions. \\
Routing and Travel & Logistics Delivery Planning & multi-stop delivery routing; capacity-aware delivery sequencing & Design delivery routes under vehicle capacity, customer time windows, and driver-hour limits. \\
Routing and Travel & Transportation Dispatch Planning & vehicle assignment for shipments; time-constrained dispatch ordering & Assign vehicles to transport jobs under fleet limits, sequencing, and deadhead minimization. \\
\end{longtable}
}
\section{Constraint Taxonomy Summary}
\label{app:constraint_taxonomy_summary}

This section gives the full appendix-level summary of the constraint taxonomy used in PlanningBench. Table~\ref{tab:appendix_general_constraints} presents the reusable general constraints shared across tasks, while Table~\ref{tab:appendix_task_specific_constraints} lists task-specific constraints for concrete planning scenarios; together they show how difficulty and verifiability are operationalized in the benchmark.

{\scriptsize
\setlength{\tabcolsep}{2.2pt}
\renewcommand{\arraystretch}{1.06}
\setlength{\LTleft}{0pt}
\setlength{\LTright}{0pt}
\setlength{\emergencystretch}{4em}
\begin{longtable}{@{}>{\RaggedRight\hspace{0pt}\arraybackslash}p{0.11\linewidth}>{\centering\arraybackslash}p{0.045\linewidth}>{\RaggedRight\hspace{0pt}\arraybackslash}p{0.17\linewidth}>{\RaggedRight\hspace{0pt}\arraybackslash}p{0.31\linewidth}>{\RaggedRight\hspace{0pt}\arraybackslash}p{0.305\linewidth}@{}}
\caption{Complete general constraints from the PlanningBench constraint summary workbook.}\label{tab:appendix_general_constraints}\\
\toprule
\textbf{Level} & \textbf{No.} & \textbf{Constraint} & \textbf{Assessed content} & \textbf{Constraint note} \\
\midrule
\endfirsthead

\toprule
\textbf{Level} & \textbf{No.} & \textbf{Constraint} & \textbf{Assessed content} & \textbf{Constraint note} \\
\midrule
\endhead

\midrule
\multicolumn{5}{r}{\scriptsize Continued on next page}\\
\endfoot

\bottomrule
\endlastfoot
Basic & 1 & Explicit primary objective & The planning task must state a clear, single, and verifiable primary objective, such as minimum conflict, minimum time, minimum cost, or maximum coverage. & Do not merely ask for a ``reasonable plan''; the optimization target must be explicit and evaluable. \\
Basic & 2 & Complete inputs & The prompt must provide key inputs, including entities, resources, time ranges, capacity bounds, and constraints. & Core parameters should not be left for the model to invent; at minimum, entities, resources, time, and boundaries should be explicit. \\
Basic & 3 & Exclusive resource usage & The same resource cannot be assigned to multiple mutually exclusive tasks at the same time. & Examples include rooms, vehicles, doctors, meeting rooms, equipment, and staff. \\
Basic & 4 & Time-window constraints & Tasks, resources, or entities may be scheduled only within the specified time windows. & Start and end times, allowed periods, or forbidden periods should be explicit, with clear time units. \\
Basic & 5 & Capacity limits & People, quantities, volume, labor hours, budgets, and similar quantities in the plan must respect hard upper bounds. & At least one hard capacity constraint should be present; over-capacity plans must be ruled infeasible. \\
Basic & 6 & Structured output & The output should use a structured format such as a table, YAML, JSON, or a time-blocked list. & Pure free-form prose is not enough; the result should be easy to inspect programmatically. \\
\midrule
Medium & 1 & Hard vs. soft constraints & Constraints should be divided into hard constraints that cannot be violated and soft constraints that should be satisfied as much as possible. & At least two hard constraints and two soft constraints should be explicit; hard constraints take precedence under conflict. \\
Medium & 2 & Multi-objective optimization & Besides the primary objective, define one to three secondary objectives for tie-breaking or solution ranking. & The priority among objectives must be explicit rather than leaving them unordered. \\
Medium & 3 & Fairness and balance & The plan should balance burdens, opportunities, durations, resource shares, or unfavorable assignments across entities as much as possible. & A concrete balance metric is expected, such as a maximum gap, variance cap, or share-deviation bound. \\
Medium & 4 & Buffer time & Reserve necessary transition, preparation, rest, commute, or cleanup time between tasks. & Schedules should not be packed edge to edge; a minimum buffer should be explicit. \\
Medium & 5 & Precedence dependencies & When prerequisite tasks, process order, non-parallel relations, or stage milestones exist, the dependency chain must be respected. & The prompt should make clear which tasks must precede others, which may run in parallel, and which may not be adjacent. \\
Medium & 6 & Conflict validation & After presenting the plan, explicitly state whether there are resource conflicts, time conflicts, capacity violations, or dependency violations. & Do not leave this implicit; a brief post-plan feasibility check is expected. \\
\midrule
Hard & 1 & Infeasibility detection & When the input conditions themselves are infeasible, the plan should identify infeasibility and point out the key source of conflict. & The system should not produce a superficially complete but actually impossible plan. \\
Hard & 2 & Minimum-change replanning & If an existing plan is provided, the revised plan should minimize the scope of changes. & Typical change costs include changed people, changed time slots, changed resources, or changed order. \\
Hard & 3 & Robustness design & The plan should remain stable under small perturbations, such as isolated resource failures, minor delays, or mild demand shifts. & This can be implemented through redundancy, backups, elastic capacity, alternate routes, or spare time slots. \\
Hard & 4 & Exception recovery strategy & Provide replanning or backup plans for common abnormal scenarios. & Typical triggers include absence, weather changes, equipment failure, delays, closures, and inventory fluctuations. \\
Hard & 5 & Local information incompleteness & When the input contains ambiguity, missing information, or uncertainty, the plan should adopt a conservative arrangement and mark any dependent assumptions. & Uncertain inputs should not be treated as certain facts; the output should flag the risk points. \\
Hard & 6 & Explainability & Give brief reasons for key arrangements to explain why they satisfy the objectives and constraints. & Explanations should focus on feasibility, optimization goals, and constraint tradeoffs rather than generic commentary. \\
\end{longtable}
}

{\scriptsize
\setlength{\tabcolsep}{2.2pt}
\renewcommand{\arraystretch}{1.06}
\setlength{\LTleft}{0pt}
\setlength{\LTright}{0pt}
\begin{longtable}{@{}>{\RaggedRight\hspace{0pt}\arraybackslash}p{0.13\linewidth}>{\RaggedRight\hspace{0pt}\arraybackslash}p{0.085\linewidth}>{\centering\arraybackslash}p{0.04\linewidth}>{\RaggedRight\hspace{0pt}\arraybackslash}p{0.17\linewidth}>{\RaggedRight\hspace{0pt}\arraybackslash}p{0.27\linewidth}>{\RaggedRight\hspace{0pt}\arraybackslash}p{0.238\linewidth}@{}}
\caption{Selected task-specific constraints from PlanningBench constraint summary workbook.}\label{tab:appendix_task_specific_constraints}\\
\toprule
\textbf{Task} & \textbf{Level} & \textbf{No.} & \textbf{Constraint} & \textbf{Assessed content} & \textbf{Constraint note} \\
\midrule
\endfirsthead

\toprule
\textbf{Task} & \textbf{Level} & \textbf{No.} & \textbf{Constraint} & \textbf{Assessed content} & \textbf{Constraint note} \\
\midrule
\endhead

\midrule
\multicolumn{6}{r}{\scriptsize Continued on next page}\\
\endfoot

\bottomrule
\endlastfoot
IT System Maintenance Planning & Basic & 1 & Maintenance Window & The schedule must respect the relevant time windows, deadlines, or temporal boundaries. & State the rule and its operational boundary explicitly. \\
IT System Maintenance Planning & Basic & 2 & Personnel Roles & Assignments must match the relevant resource, role, qualification, skill, capability, or venue requirements. & State this explicitly and keep the plan feasible under it. \\
IT System Maintenance Planning & Basic & 3 & Change Order & The schedule must respect the relevant precedence relations, process order, milestones, or dependency chain. & State this explicitly and keep the plan feasible under it. \\
IT System Maintenance Planning & Basic & 4 & Service-Impact Boundary & The plan must stay within the specified quantitative limits, capacity bounds, or budget boundaries. & Model the requirement directly rather than satisfying it only superficially. \\
IT System Maintenance Planning & Basic & 5 & System Dependency Topology & The schedule must respect the relevant precedence relations, process order, milestones, or dependency chain. & State this explicitly and keep the plan feasible under it. \\
IT System Maintenance Planning & Basic & 6 & Maintenance Window Length & The schedule must respect the relevant time windows, deadlines, or temporal boundaries. & State this explicitly and keep the plan feasible under it. \\
IT System Maintenance Planning & Basic & 7 & Rollback Conditions & The plan should support explicit local recovery, substitution, rollback, or replanning when conditions change. & State this explicitly and keep the plan feasible under it. \\
IT System Maintenance Planning & Basic & 8 & Complete On-Call Role Coverage & Assignments must match the relevant resource, role, qualification, skill, capability, or venue requirements. & State this explicitly and keep the plan feasible under it. \\
\midrule
IT System Maintenance Planning & Medium & 1 & Canary or Phased Strategy & The plan should encode and enforce the required priority, protection, or triage logic. & Use this to refine solution quality beyond basic feasibility. \\
IT System Maintenance Planning & Medium & 2 & Validation Checkpoints & The plan should encode and enforce the required priority, protection, or triage logic. & Model the requirement directly rather than satisfying it only superficially. \\
IT System Maintenance Planning & Medium & 3 & Resource Coordination & The plan should coordinate the relevant people, resources, or stages explicitly rather than assuming they line up automatically. & Use this to refine solution quality beyond basic feasibility. \\
IT System Maintenance Planning & Medium & 4 & Rollback Plan & The plan should support explicit local recovery, substitution, rollback, or replanning when conditions change. & At least one concrete fallback should be available. \\
IT System Maintenance Planning & Medium & 5 & Canary Rollout Order & The schedule must respect the relevant precedence relations, process order, milestones, or dependency chain. & Use this to refine solution quality beyond basic feasibility. \\
IT System Maintenance Planning & Medium & 6 & Prioritize Business Off-Peak Windows & The schedule must respect the relevant time windows, deadlines, or temporal boundaries. & Use this to refine solution quality beyond basic feasibility. \\
IT System Maintenance Planning & Medium & 7 & Monitoring Observation Window & The schedule must respect the relevant time windows, deadlines, durations, or temporal boundaries. & Use this to refine solution quality beyond basic feasibility. \\
IT System Maintenance Planning & Medium & 8 & Cross-Team Coordination Load & The plan should coordinate the relevant people, resources, or stages explicitly rather than assuming they line up automatically. & Use this to refine solution quality beyond basic feasibility. \\
\midrule
IT System Maintenance Planning & Hard & 1 & Failure-to-Recovery Switching & The plan should support explicit local recovery, substitution, rollback, or replanning when conditions change. & Model the requirement directly rather than satisfying it only superficially. \\
IT System Maintenance Planning & Hard & 2 & Minimum Business Disruption & The plan should support explicit local recovery, substitution, rollback, or replanning when conditions change. & Prefer local repair and minimum disruption rather than rebuilding the full plan. \\
IT System Maintenance Planning & Hard & 3 & Infeasibility Identification & When the requirement cannot be met, the plan should diagnose infeasibility and identify the main bottleneck. & Model the requirement directly rather than satisfying it only superficially. \\
IT System Maintenance Planning & Hard & 4 & Multi-Objective Comparison & The plan should support explicit local recovery, substitution, rollback, or replanning when conditions change. & Make the recovery, tradeoff, or diagnosis logic explicit. \\
IT System Maintenance Planning & Hard & 5 & Emergency Fault Insertion & The plan should encode and enforce the required priority, protection, or triage logic. & The priority rule should have a visible effect on the final ordering or allocation. \\
IT System Maintenance Planning & Hard & 6 & Version-Failure Rollback & The plan should support explicit local recovery, substitution, rollback, or replanning when conditions change. & Make the recovery, tradeoff, or diagnosis logic explicit. \\
IT System Maintenance Planning & Hard & 7 & Coupled Multi-System Failure & The schedule must respect the relevant precedence relations, process order, or dependency chain. & Make the recovery, tradeoff, or diagnosis logic explicit. \\
IT System Maintenance Planning & Hard & 8 & Cross-Region Consistency Recovery & The plan should support explicit local recovery, substitution, rollback, or replanning when conditions change. & State the rule and its operational boundary explicitly. \\
\midrule
Meeting Planning & Basic & 1 & Key Attendee Availability & Only feasible and available people, resources, locations, routes, or periods may be used. & The priority rule should have a visible effect on the final ordering or allocation. \\
Meeting Planning & Basic & 2 & Meeting Room Capacity and Equipment Match & The plan must stay within the specified quantitative limits, capacity bounds, or budget boundaries. & State this explicitly and keep the plan feasible under it. \\
Meeting Planning & Basic & 3 & Valid Duration & The schedule must respect the relevant time windows, deadlines, durations, or temporal boundaries. & State this explicitly and keep the plan feasible under it. \\
Meeting Planning & Basic & 4 & No Time Conflict & The schedule must respect the relevant time windows, deadlines, or temporal boundaries. & Treat this as a hard feasibility constraint. \\
Meeting Planning & Basic & 5 & Complete Attendance Role Coverage & Assignments must match the relevant resource, role, qualification, skill, capability, or venue requirements. & State this explicitly and keep the plan feasible under it. \\
Meeting Planning & Basic & 6 & Equipment-Demand Matching & Assignments must match the relevant resource, role, qualification, skill, capability, or venue requirements. & State this explicitly and keep the plan feasible under it. \\
Meeting Planning & Basic & 7 & Minimum Meeting Duration & The schedule must respect the relevant time windows, deadlines, durations, or temporal boundaries. & Prefer local repair and minimum disruption rather than rebuilding the full plan. \\
Meeting Planning & Basic & 8 & Daily Meeting Count Limit & The plan must stay within the specified quantitative limits, capacity bounds, or budget boundaries. & State this explicitly and keep the plan feasible under it. \\
\midrule
Meeting Planning & Medium & 1 & Reduce Schedule Fragmentation & The plan should explicitly model the requirement captured by this reduce schedule fragmentation. & Use this to refine solution quality beyond basic feasibility. \\
Meeting Planning & Medium & 2 & Cross-Time-Zone Fairness & The plan should keep the relevant loads, opportunities, or burdens reasonably balanced. & Use this to refine solution quality beyond basic feasibility. \\
Meeting Planning & Medium & 3 & Meeting Priority & The plan should encode and enforce the required priority, protection, or triage logic. & State the rule and its operational boundary explicitly. \\
Meeting Planning & Medium & 4 & Consecutive-Meeting Control & The plan must stay within the specified quantitative limits and capacity boundaries. & Use this to refine solution quality beyond basic feasibility. \\
Meeting Planning & Medium & 5 & Time-Zone Fairness & The plan should keep the relevant loads, opportunities, or burdens reasonably balanced. & Use this to refine solution quality beyond basic feasibility. \\
Meeting Planning & Medium & 6 & Consecutive-Meeting Buffer & The plan should include an explicit buffer, backup, contingency, or reserve mechanism. & Use this to refine solution quality beyond basic feasibility. \\
Meeting Planning & Medium & 7 & Priority Ranking Rule & The plan should encode and enforce the required priority, protection, or triage logic. & State the rule and its operational boundary explicitly. \\
Meeting Planning & Medium & 8 & Meeting-Chain Order & The schedule must respect the relevant precedence relations, process order, milestones, or dependency chain. & Use this to refine solution quality beyond basic feasibility. \\
\midrule
Meeting Planning & Hard & 1 & Minimum Disruption from Rescheduling & The plan should support explicit local recovery, substitution, rollback, or replanning when conditions change. & Prefer local repair and minimum disruption rather than rebuilding the full plan. \\
Meeting Planning & Hard & 2 & Backup Time Slots & The plan should include an explicit buffer, backup, contingency, or reserve mechanism. & At least one concrete fallback should be available. \\
Meeting Planning & Hard & 3 & Local Information Incompleteness & The plan must satisfy the required completeness or minimum-service condition. & Make the recovery, tradeoff, or diagnosis logic explicit. \\
Meeting Planning & Hard & 4 & Global Coordination across Multiple Meetings & The plan should coordinate the relevant people, resources, or stages explicitly rather than assuming they line up automatically. & Make the recovery, tradeoff, or diagnosis logic explicit. \\
Meeting Planning & Hard & 5 & Recovery from Key-Person Changes & The plan should support explicit local recovery, substitution, rollback, or replanning when conditions change. & The priority rule should have a visible effect on the final ordering or allocation. \\
Meeting Planning & Hard & 6 & Temporary Meeting-Room Failure & The plan should support explicit local recovery, substitution, rollback, or replanning when conditions change. & Model the requirement directly rather than satisfying it only superficially. \\
Meeting Planning & Hard & 7 & Agenda Splitting and Merging & The schedule must respect the relevant time windows, deadlines, or temporal boundaries. & State the rule and its operational boundary explicitly. \\
Meeting Planning & Hard & 8 & Confidentiality-Level Layering & Assignments must match the relevant resource, role, qualification, capability, or venue requirements. & Make the recovery, tradeoff, or diagnosis logic explicit. \\
\midrule
Medical Resource Planning & Basic & 1 & Triage by Urgency and Severity & The plan should encode and enforce the required priority, protection, or triage logic. & The priority rule should have a visible effect on the final ordering or allocation. \\
Medical Resource Planning & Basic & 2 & Doctor-Room Matching & Assignments must match the relevant resource, role, qualification, skill, capability, or venue requirements. & Model the requirement directly rather than satisfying it only superficially. \\
Medical Resource Planning & Basic & 3 & Resource Capacity & The plan must stay within the specified quantitative limits, capacity bounds, or budget boundaries. & Treat this as a hard feasibility constraint. \\
Medical Resource Planning & Basic & 4 & Time Availability & Only feasible and available people, resources, locations, routes, or periods may be used. & State this explicitly and keep the plan feasible under it. \\
Medical Resource Planning & Basic & 5 & Emergency Priority & The plan should encode and enforce the required priority, protection, or triage logic. & The priority rule should have a visible effect on the final ordering or allocation. \\
Medical Resource Planning & Basic & 6 & Doctor Specialty Matching & Assignments must match the relevant resource, role, qualification, skill, capability, or venue requirements. & State this explicitly and keep the plan feasible under it. \\
Medical Resource Planning & Basic & 7 & Room-Function Matching & Assignments must match the relevant resource, role, qualification, skill, capability, or venue requirements. & State this explicitly and keep the plan feasible under it. \\
Medical Resource Planning & Basic & 8 & Waiting-Area Capacity & The plan must stay within the specified quantitative limits, capacity bounds, or budget boundaries. & State this explicitly and keep the plan feasible under it. \\
\midrule
Medical Resource Planning & Medium & 1 & Waiting-Time Control & The plan should explicitly model the requirement captured by this waiting-time control. & Use this to refine solution quality beyond basic feasibility. \\
Medical Resource Planning & Medium & 2 & Fairness & The plan should keep the relevant loads, opportunities, or burdens reasonably balanced. & Use this to refine solution quality beyond basic feasibility. \\
Medical Resource Planning & Medium & 3 & Peak-Time Elasticity & The schedule must respect the relevant time windows, deadlines, or temporal boundaries. & Use this to improve robustness rather than only nominal feasibility. \\
Medical Resource Planning & Medium & 4 & Continuity across Care Stages & The schedule must respect the relevant precedence relations, process order, milestones, or dependency chain. & Use this to refine solution quality beyond basic feasibility. \\
Medical Resource Planning & Medium & 5 & Waiting-Time Balance & The plan should keep the relevant loads, opportunities, or burdens reasonably balanced. & The priority rule should have a visible effect on the final ordering or allocation. \\
Medical Resource Planning & Medium & 6 & Follow-Up Coordination & The plan should coordinate the relevant people, resources, or stages explicitly rather than assuming they line up automatically. & Use this to refine solution quality beyond basic feasibility. \\
Medical Resource Planning & Medium & 7 & Doctor Workload Balance & The plan must stay within the specified quantitative limits, capacity bounds, or budget boundaries. & Use this to refine solution quality beyond basic feasibility. \\
Medical Resource Planning & Medium & 8 & Equipment Utilization & The plan must stay within the specified quantitative limits, capacity bounds, or budget boundaries. & Use this to refine solution quality beyond basic feasibility. \\
\midrule
Medical Resource Planning & Hard & 1 & Recovery from Sudden Patient Surge & The plan should support explicit local recovery, substitution, rollback, or replanning when conditions change. & The priority rule should have a visible effect on the final ordering or allocation. \\
Medical Resource Planning & Hard & 2 & Rescheduling after Doctor Absence & The plan should support explicit local recovery, substitution, rollback, or replanning when conditions change. & Make the recovery, tradeoff, or diagnosis logic explicit. \\
Medical Resource Planning & Hard & 3 & Infeasibility Diagnosis & When the requirement cannot be met, the plan should diagnose infeasibility and identify the main bottleneck. & Model the requirement directly rather than satisfying it only superficially. \\
Medical Resource Planning & Hard & 4 & Multi-Objective Comparison & The plan should encode and enforce the required priority, protection, or triage logic. & Make the recovery, tradeoff, or diagnosis logic explicit. \\
Medical Resource Planning & Hard & 5 & Recovery from Doctor Service Suspension & The plan should support explicit local recovery, substitution, rollback, or replanning when conditions change. & The priority rule should have a visible effect on the final ordering or allocation. \\
Medical Resource Planning & Hard & 6 & Inspection-Delay Cascade & The plan should explicitly model the requirement captured by this inspection-delay cascade. & Model the requirement directly rather than satisfying it only superficially. \\
Medical Resource Planning & Hard & 7 & Bed/Room Bottleneck Identification & When the requirement cannot be met, the plan should diagnose infeasibility and identify the main bottleneck. & Make the recovery, tradeoff, or diagnosis logic explicit. \\
Medical Resource Planning & Hard & 8 & Handling Multi-Priority Conflicts & The plan should encode and enforce the required priority, protection, or triage logic. & The priority rule should have a visible effect on the final ordering or allocation. \\
\midrule
Work Assignment Scheduling & Basic & 1 & Task-Person Matching & Assignments must match the relevant resource, role, qualification, skill, capability, or venue requirements. & State this explicitly and keep the plan feasible under it. \\
Work Assignment Scheduling & Basic & 2 & Deadline Satisfaction & The schedule must respect the relevant time windows, deadlines, durations, or temporal boundaries. & State this explicitly and keep the plan feasible under it. \\
Work Assignment Scheduling & Basic & 3 & Dependencies & The schedule must respect the relevant precedence relations, process order, or dependency chain. & State the rule and its operational boundary explicitly. \\
Work Assignment Scheduling & Basic & 4 & Basic Balance & The plan should keep the relevant loads, opportunities, or burdens reasonably balanced. & State this explicitly and keep the plan feasible under it. \\
Work Assignment Scheduling & Basic & 5 & Task-Skill Matching & Assignments must match the relevant resource, role, qualification, skill, capability, or venue requirements. & State this explicitly and keep the plan feasible under it. \\
Work Assignment Scheduling & Basic & 6 & Availability of Time Estimates & Only feasible and available people, resources, locations, routes, or periods may be used. & State this explicitly and keep the plan feasible under it. \\
Work Assignment Scheduling & Basic & 7 & Task Non-Parallelizability & The plan should explicitly model the requirement captured by this task non-parallelizability. & State this explicitly and keep the plan feasible under it. \\
Work Assignment Scheduling & Basic & 8 & Explicit Delivery Milestones & The schedule must respect the relevant precedence relations, process order, milestones, or dependency chain. & Model the requirement directly rather than satisfying it only superficially. \\
\midrule
Work Assignment Scheduling & Medium & 1 & Use of Parallelism & The plan should explicitly model the requirement captured by this use of parallelism. & Use this to refine solution quality beyond basic feasibility. \\
Work Assignment Scheduling & Medium & 2 & Priority Queue & The plan should encode and enforce the required priority, protection, or triage logic. & State the rule and its operational boundary explicitly. \\
Work Assignment Scheduling & Medium & 3 & Handover Cost & The plan should coordinate the relevant people, resources, or stages explicitly rather than assuming they line up automatically. & Use this to refine solution quality beyond basic feasibility. \\
Work Assignment Scheduling & Medium & 4 & Stage Checkpoints & The plan should explicitly model the requirement captured by this stage checkpoints. & Use this to refine solution quality beyond basic feasibility. \\
Work Assignment Scheduling & Medium & 5 & Individual Workload Balance & The plan must stay within the specified quantitative limits, capacity bounds, or budget boundaries. & Use this to refine solution quality beyond basic feasibility. \\
Work Assignment Scheduling & Medium & 6 & Collaboration Handover Buffer & The plan should coordinate the relevant people, resources, or stages explicitly rather than assuming they line up automatically. & Use this to refine solution quality beyond basic feasibility. \\
Work Assignment Scheduling & Medium & 7 & Protection of Key Personnel & The plan should encode and enforce the required priority, protection, or triage logic. & Use this to refine solution quality beyond basic feasibility. \\
Work Assignment Scheduling & Medium & 8 & Interleaving High- and Low-Priority Work & The plan should encode and enforce the required priority, protection, or triage logic. & The priority rule should have a visible effect on the final ordering or allocation. \\
\midrule
Work Assignment Scheduling & Hard & 1 & Minimum Adjustment to Existing Plan & The plan should support explicit local recovery, substitution, rollback, or replanning when conditions change. & Prefer local repair and minimum disruption rather than rebuilding the full plan. \\
Work Assignment Scheduling & Hard & 2 & Recovery from Key-Member Absence & The plan should support explicit local recovery, substitution, rollback, or replanning when conditions change. & Make the recovery, tradeoff, or diagnosis logic explicit. \\
Work Assignment Scheduling & Hard & 3 & Dual Objective: Schedule and Cost & The plan should explicitly model the requirement captured by this dual objective: schedule and cost. & State the rule and its operational boundary explicitly. \\
Work Assignment Scheduling & Hard & 4 & Infeasibility Bottleneck Diagnosis & When the requirement cannot be met, the plan should diagnose infeasibility and identify the main bottleneck. & Make the diagnosis logic explicit and identify what blocks feasibility. \\
Work Assignment Scheduling & Hard & 5 & Rescheduling after Demand Insertion & The plan should support explicit local recovery, substitution, rollback, or replanning when conditions change. & The priority rule should have a visible effect on the final ordering or allocation. \\
Work Assignment Scheduling & Hard & 6 & Recovery from Member Leave & The plan should support explicit local recovery, substitution, rollback, or replanning when conditions change. & Make the recovery, tradeoff, or diagnosis logic explicit. \\
Work Assignment Scheduling & Hard & 7 & Bottleneck Task Identification & When the requirement cannot be met, the plan should diagnose infeasibility and identify the main bottleneck. & Model the requirement directly rather than satisfying it only superficially. \\
Work Assignment Scheduling & Hard & 8 & Cross-Week Continuity & The plan should explicitly model the requirement captured by this cross-week continuity. & Make the recovery, tradeoff, or diagnosis logic explicit. \\
\midrule
Logistics Delivery Planning & Basic & 1 & Vehicle Capacity & The plan must stay within the specified quantitative limits, capacity bounds, or budget boundaries. & State this explicitly and keep the plan feasible under it. \\
Logistics Delivery Planning & Basic & 2 & Customer Time Windows & The schedule must respect the relevant time windows, deadlines, durations, or temporal boundaries. & State this explicitly and keep the plan feasible under it. \\
Logistics Delivery Planning & Basic & 3 & Route Feasibility & Only feasible and available people, resources, locations, routes, or periods may be used. & Model the requirement directly rather than satisfying it only superficially. \\
Logistics Delivery Planning & Basic & 4 & Exclusive Vehicle Usage & The plan should explicitly model the requirement captured by this exclusive vehicle usage. & Treat this as a hard feasibility constraint. \\
Logistics Delivery Planning & Basic & 5 & Order Time Windows & The schedule must respect the relevant time windows, deadlines, durations, or temporal boundaries. & State this explicitly and keep the plan feasible under it. \\
Logistics Delivery Planning & Basic & 6 & Vehicle Capacity Dimensions & The plan must stay within the specified quantitative limits, capacity bounds, or budget boundaries. & State the rule and its operational boundary explicitly. \\
Logistics Delivery Planning & Basic & 7 & Driver Working-Hour Limits & The plan must stay within the specified quantitative limits, capacity bounds, or budget boundaries. & State this explicitly and keep the plan feasible under it. \\
Logistics Delivery Planning & Basic & 8 & Loading and Unloading Time & The plan must stay within the specified quantitative limits, capacity bounds, or budget boundaries. & Model the requirement directly rather than satisfying it only superficially. \\
\midrule
Logistics Delivery Planning & Medium & 1 & Regional Partitioning & The plan should explicitly model the requirement captured by this regional partitioning. & Use this to refine solution quality beyond basic feasibility. \\
Logistics Delivery Planning & Medium & 2 & Priority Order Protection & The schedule must respect the relevant precedence relations, process order, milestones, or dependency chain. & State the rule and its operational boundary explicitly. \\
Logistics Delivery Planning & Medium & 3 & Return-to-Depot Strategy & The plan should explicitly model the requirement captured by this return-to-depot strategy. & Use this to refine solution quality beyond basic feasibility. \\
Logistics Delivery Planning & Medium & 4 & Driver Working Hours & The plan must stay within the specified quantitative limits, capacity bounds, or budget boundaries. & Use this to refine solution quality beyond basic feasibility. \\
Logistics Delivery Planning & Medium & 5 & Regional Batching & The plan should explicitly model the requirement captured by this regional batching. & Use this to refine solution quality beyond basic feasibility. \\
Logistics Delivery Planning & Medium & 6 & High-Priority Orders First & The schedule must respect the relevant precedence relations, process order, milestones, or dependency chain. & The priority rule should have a visible effect on the final ordering or allocation. \\
Logistics Delivery Planning & Medium & 7 & Cold-Chain / Regular Mixed-Load Restrictions & The plan must stay within the specified quantitative limits, capacity bounds, or budget boundaries. & Use this to refine solution quality beyond basic feasibility. \\
Logistics Delivery Planning & Medium & 8 & Return-Trip Utilization & The plan must stay within the specified quantitative limits, capacity bounds, or budget boundaries. & Use this to refine solution quality beyond basic feasibility. \\
\midrule
Logistics Delivery Planning & Hard & 1 & Recovery from Traffic Disruption & The plan should support explicit local recovery, substitution, rollback, or replanning when conditions change. & At least one concrete fallback should be available. \\
Logistics Delivery Planning & Hard & 2 & Minimum Disruption from Inserted Orders & The plan should support explicit local recovery, substitution, rollback, or replanning when conditions change. & Prefer local repair and minimum disruption rather than rebuilding the full plan. \\
Logistics Delivery Planning & Hard & 3 & Multi-Objective Comparison & The plan should remain workable under uncertainty, mild disturbance, or modest parameter variation. & Model the requirement directly rather than satisfying it only superficially. \\
Logistics Delivery Planning & Hard & 4 & Infeasibility Diagnosis & When the requirement cannot be met, the plan should diagnose infeasibility and identify the main bottleneck. & Make the diagnosis logic explicit and identify what blocks feasibility. \\
Logistics Delivery Planning & Hard & 5 & Insertion of New Orders & The schedule must respect the relevant precedence relations, process order, milestones, or dependency chain. & Make the recovery, tradeoff, or diagnosis logic explicit. \\
Logistics Delivery Planning & Hard & 6 & Recovery from Road Closure & The plan should support explicit local recovery, substitution, rollback, or replanning when conditions change. & Make the recovery, tradeoff, or diagnosis logic explicit. \\
Logistics Delivery Planning & Hard & 7 & Substitution after Vehicle Failure & The plan should support explicit local recovery, substitution, rollback, or replanning when conditions change. & Make the recovery, tradeoff, or diagnosis logic explicit. \\
Logistics Delivery Planning & Hard & 8 & Multi-Warehouse Cooperative Backfill & The plan should support explicit local recovery, substitution, rollback, or replanning when conditions change. & Make the recovery, tradeoff, or diagnosis logic explicit. \\
\midrule
Material Allocation & Basic & 1 & Total Quantity Constraint & The plan must stay within the specified quantitative limits, capacity bounds, or budget boundaries. & Treat this as a hard feasibility constraint. \\
Material Allocation & Basic & 2 & Demand Matching & Assignments must match the relevant resource, role, qualification, skill, capability, or venue requirements. & State this explicitly and keep the plan feasible under it. \\
Material Allocation & Basic & 3 & Unique Counting & The plan should explicitly model the requirement captured by this unique counting. & State this explicitly and keep the plan feasible under it. \\
Material Allocation & Basic & 4 & Minimum Guarantee & The plan must stay within the specified quantitative limits and capacity boundaries. & Prefer local repair and minimum disruption rather than rebuilding the full plan. \\
Material Allocation & Basic & 5 & Minimum Guarantee Threshold & The plan must stay within the specified quantitative limits, capacity bounds, or budget boundaries. & Prefer local repair and minimum disruption rather than rebuilding the full plan. \\
Material Allocation & Basic & 6 & Demand Validity & Assignments must match the relevant resource, role, qualification, capability, or venue requirements. & State this explicitly and keep the plan feasible under it. \\
Material Allocation & Basic & 7 & Allocation Granularity & The plan should explicitly model the requirement captured by this allocation granularity. & State this explicitly and keep the plan feasible under it. \\
Material Allocation & Basic & 8 & Use-Case Matching & Assignments must match the relevant resource, role, qualification, skill, capability, or venue requirements. & State this explicitly and keep the plan feasible under it. \\
\midrule
Material Allocation & Medium & 1 & Tiered Prioritization & Assignments must match the relevant resource, role, qualification, capability, or venue requirements. & State the rule and its operational boundary explicitly. \\
Material Allocation & Medium & 2 & Fairness Boundary & The plan must stay within the specified quantitative limits, capacity bounds, or budget boundaries. & Use this to refine solution quality beyond basic feasibility. \\
Material Allocation & Medium & 3 & Secondary Allocation Mechanism & The plan should explicitly model the requirement captured by this secondary allocation mechanism. & Use this to refine solution quality beyond basic feasibility. \\
Material Allocation & Medium & 4 & Adapt to Demand Updates & Assignments must match the relevant resource, role, qualification, capability, or venue requirements. & Use this to refine solution quality beyond basic feasibility. \\
Material Allocation & Medium & 5 & Priority Tilt Magnitude & The plan should encode and enforce the required priority, protection, or triage logic. & The priority rule should have a visible effect on the final ordering or allocation. \\
Material Allocation & Medium & 6 & Historical Fairness Compensation & The plan should keep the relevant loads, opportunities, or burdens reasonably balanced. & Use this to refine solution quality beyond basic feasibility. \\
Material Allocation & Medium & 7 & Timeliness Priority & The plan should encode and enforce the required priority, protection, or triage logic. & State the rule and its operational boundary explicitly. \\
Material Allocation & Medium & 8 & Strategy for Staged Arrivals & The schedule must respect the relevant time windows, deadlines, durations, or temporal boundaries. & Use this to refine solution quality beyond basic feasibility. \\
\midrule
Material Allocation & Hard & 1 & Temporary New High-Priority Demand & The plan should encode and enforce the required priority, protection, or triage logic. & The priority rule should have a visible effect on the final ordering or allocation. \\
Material Allocation & Hard & 2 & Multi-Objective Comparison & The plan should keep the relevant loads, opportunities, or burdens reasonably balanced. & Make the recovery, tradeoff, or diagnosis logic explicit. \\
Material Allocation & Hard & 3 & Infeasibility Analysis & When the requirement cannot be met, the plan should diagnose infeasibility and identify the main bottleneck. & Make the diagnosis logic explicit and identify what blocks feasibility. \\
Material Allocation & Hard & 4 & Robustness & The plan should remain workable under uncertainty, safety constraints, or modest operational disturbance. & Use this to improve robustness rather than only nominal feasibility. \\
Material Allocation & Hard & 5 & Absorb New Demanders & The plan should support explicit local recovery, substitution, rollback, or replanning when conditions change. & Make the recovery, tradeoff, or diagnosis logic explicit. \\
Material Allocation & Hard & 6 & Infeasibility Detection under Demand Surge & When the requirement cannot be met, the plan should diagnose infeasibility and identify the main bottleneck. & Make the diagnosis logic explicit and identify what blocks feasibility. \\
Material Allocation & Hard & 7 & Error Buffer & The plan should include an explicit buffer, backup, contingency, or reserve mechanism. & Use this to improve robustness rather than only nominal feasibility. \\
Material Allocation & Hard & 8 & Cross-Region Reallocation & The plan should explicitly model the requirement captured by this cross-region reallocation. & Make the recovery, tradeoff, or diagnosis logic explicit. \\
\end{longtable}
}
\section{Implementation Details for Constraint Sampling}
\label{app:constraint_sampling}

This appendix provides additional details on constraint-count sampling, subset construction, and projection in the synthesis pipeline. These details are kept in the appendix because the main text focuses on the high-level closed-loop synthesis procedure.

\paragraph{Initial constraint-count sampling.}
For each sampled task \(\tau\) and subtask variant \(\sigma\), the Generator samples constraints from the task-specific basic, medium, and hard constraint pools. The initial number of sampled constraints from each pool is drawn from a fixed categorical prior:
\[
N_b \sim \{1\!:\!0.2,2\!:\!0.6,3\!:\!0.2\},
\quad
N_m \sim \{0\!:\!0.25,1\!:\!0.55,2\!:\!0.2\},
\quad
N_h \sim \{0\!:\!0.7,1\!:\!0.3\}.
\]
These priors ensure that each instance contains at least one basic constraint, usually includes medium-level requirements, and introduces hard constraints sparsely. This design keeps early generated instances manageable while allowing closed-loop difficulty enhancement to shift later samples toward harder constraint compositions.

\paragraph{Subset construction.}
After drawing \(N_b\), \(N_m\), and \(N_h\), the Generator samples without replacement
\[
S_b \subset \mathcal{C}_b(\tau),\qquad
S_m \subset \mathcal{C}_m(\tau),\qquad
S_h \subset \mathcal{C}_h(\tau),
\]
with \(|S_b|=N_b\), \(|S_m|=N_m\), and \(|S_h|=N_h\). These sampled subsets are checked against task-template compatibility rules to avoid duplicate, contradictory, or underspecified requirements. Specialized stateful constraints are maintained as an optional layer and are sampled separately when the corresponding task template requires state-dependent behavior.

Together with a random variable \(z\) controlling wording, numerical instantiation, and contextual details, the sampled elements define the generation specification
\[
r=(\tau,\sigma,S_b,S_m,S_h;z).
\]
The Generator then produces a problem and checklist pair \((x,c)=G(r)\).

\paragraph{Projection after difficulty resampling.}
After adaptive difficulty resampling updates the difficulty-level probabilities, the updated distribution is converted back into discrete constraint counts. Given the next-turn total constraint number \(M^{(k+1)}\), we compute the expected allocation
\[
\mathbb{E}[\mathbf{n}^{(k+1)}]=M^{(k+1)}\mathbf{p}^{(k+1)}.
\]
This expected allocation is projected onto the admissible count space
\[
\mathcal{S}=\{1,2,3\}\times\{0,1,2\}\times\{0,1\}
\]
through \(\Pi_{\mathcal{S}}(\cdot)\), yielding
\[
(N_b^{(k+1)},N_m^{(k+1)},N_h^{(k+1)}).
\]
This projection preserves the allowed count ranges for basic, medium, and hard constraints while reflecting the updated difficulty distribution. It also keeps the problem size controllable during adaptive resampling.
\section{Human Quality-control Categories}
\label{app:quality_control}

During human quality control, annotators assign each synthesized sample to one of four categories:
\begin{enumerate}[label=(\arabic*),leftmargin=1.7em,itemsep=0.12em,topsep=0.3em]
    \item \emph{No modification needed}. The prompt, checklist, and reference answer are clear, consistent, and directly usable.
    \item \emph{Minor revision with usable source data}. The instance is generally valid, but minor edits are needed to improve wording, formatting, checklist coverage, or answer clarity.
    \item \emph{Minor revision with source data not directly usable}. The instance can be retained only after correcting or completing source information, such as missing rules, incomplete boundary conditions, or underspecified constraints.
    \item \emph{Discard}. The instance contains irrecoverable ambiguity, inconsistency, or verification failure and is removed from the final data.
\end{enumerate}

Across the audited batch, 86.15\% of synthesized samples fall into categories (1) and (2), while 13.85\% require category-(3) revision. No sample in this batch is directly assigned to the discard category. Most revisions involve missing checklist items, incomplete constraint checks, ambiguous or underspecified prompts, unclear references, failures to follow required output structures, missing rules or boundary conditions, and occasional logical inconsistencies between the prompt and answer.
\section{Break Case}
\label{app:break_case}

This section provides a concrete break case used for qualitative inspection of model failures. Table~\ref{tab:break_case_production_planning} shows a production-planning example together with the gold solution, a model response, and the corresponding error analysis, illustrating the kinds of global-consistency mistakes that are penalized by PlanningBench.

\begingroup
\setlength{\parskip}{0pt}
\setlength{\LTleft}{0pt}
\setlength{\LTright}{0pt}
\setlength{\tabcolsep}{0pt}
\renewcommand{\arraystretch}{1.08}
\setlength{\abovedisplayskip}{4pt}
\setlength{\belowdisplayskip}{4pt}
\setlength{\abovedisplayshortskip}{2pt}
\setlength{\belowdisplayshortskip}{2pt}
\setlist[itemize]{leftmargin=1.15em,itemsep=1pt,topsep=1pt,parsep=0pt,partopsep=0pt}
\setlist[enumerate]{leftmargin=1.25em,itemsep=1pt,topsep=1pt,parsep=0pt,partopsep=0pt}

\begin{longtable}{@{}p{\linewidth}@{}}
\caption{Representative break case from PlanningBench.}
\label{tab:break_case_production_planning}\\
\toprule
\textbf{Category}: Project \& Production Operations \hfill \textbf{ID}: D1-Day \\
\endfirsthead

\caption[]{Representative break case from PlanningBench (continued).}\\
\toprule
\textbf{Category}: Project \& Production Operations \hfill \textbf{ID}: D1-Day \\
\midrule
\endhead

\bottomrule
\endfoot

\midrule
\textbf{Task} \\

The planner is responsible for the \textbf{L-01 single-line assembly workshop} over the next \textbf{three days}, with only \textbf{two shifts per day}: a \textbf{day shift} and a \textbf{night shift}. The requested output is a \textbf{concrete production and delivery plan} that can be directly issued to the workshop, rather than high-level advice. \\

There is only \textbf{one production line}, and \textbf{all orders must be processed serially}. Orders may be split \textbf{only by whole batches across different shifts}; \textbf{no batch may be split in half}. Each batch always takes \textbf{30 minutes}. \textbf{Overtime is not allowed}, \textbf{no temporary shifts may be added}, and \textbf{no other equipment may be borrowed}. \\

The \textbf{theoretical capacity} of each shift is \textbf{480 minutes = 16 batches}, but each shift has fixed overheads of \textbf{60 minutes for equipment inspection} and \textbf{60 minutes for line clearance / first-piece confirmation}. Therefore, the \textbf{effective schedulable capacity} is only \textbf{360 minutes = 12 batches}. Scheduling must use \textbf{effective capacity} rather than theoretical capacity. \\

The six shifts, in order, are \textbf{D1-Day, D1-Night, D2-Day, D2-Night, D3-Day, D3-Night}. \\

The main raw material \textbf{R} arrives only before the start of the day shift: \\
\begin{itemize}
    \item before \textbf{D1-Day}: \textbf{188 kg}
    \item before \textbf{D2-Day}: another \textbf{194 kg}
    \item before \textbf{D3-Day}: another \textbf{174 kg}
\end{itemize}
\\

No arrivals occur during a shift, and material that has not yet arrived cannot be pre-allocated. An order becomes deliverable only after the \textbf{entire order} is completed. The due date for each order is \textbf{before the end of the night shift on its due day}. \\

There is also a shop-floor rule: \textbf{within the same shift, if the current order can still continue, do not switch back and forth between orders merely to make the schedule look balanced}. Across shifts, however, the next order should be selected again according to the target priority. \\

The order set is: \\
\begin{itemize}
    \item \textbf{O1}: 10 batches, 30 min/batch, consumes \textbf{8 kg R/batch}, due on \textbf{D1}
    \item \textbf{O2}: 8 batches, 30 min/batch, consumes \textbf{10 kg R/batch}, due on \textbf{D1}
    \item \textbf{O3}: 12 batches, 30 min/batch, consumes \textbf{9 kg R/batch}, due on \textbf{D2}
    \item \textbf{O4}: 15 batches, 30 min/batch, consumes \textbf{7 kg R/batch}, due on \textbf{D2}
    \item \textbf{O5}: 11 batches, 30 min/batch, consumes \textbf{9 kg R/batch}, due on \textbf{D3}
    \item \textbf{O6}: 14 batches, 30 min/batch, consumes \textbf{6 kg R/batch}, due on \textbf{D3}
\end{itemize}
\\

The response must explicitly follow this objective hierarchy: \\
\begin{itemize}
    \item \textbf{Primary objective}: minimize the total number of unfinished batches across all orders at their due dates
    \item \textbf{Secondary objective}: subject to the primary optimum, minimize the maximum load gap across the six shifts
    \item \textbf{Tie-breaking}: earlier due date first; for the same due date, ascending order ID
\end{itemize}
\\

The required answer format is: \\
\begin{itemize}
    \item \textbf{A.} Primary objective / Secondary objective / Tie-breaking rule
    \item \textbf{B.} Shift-by-shift production schedule
    \item \textbf{C.} Order delivery results
    \item \textbf{D.} Constraint check
\end{itemize}
\\[2pt]

\midrule
\textbf{Rubrics} \\

A correct response should be equivalent in meaning to the following plan. \\

It should explicitly state: \\
\begin{itemize}
    \item \textbf{Primary objective}: minimize unfinished batches at due dates, equivalently achieving zero lateness
    \item \textbf{Secondary objective}: subject to the primary optimum, minimize the maximum difference in utilized load across the six shifts
    \item each shift has \textbf{theoretical capacity = 16 batches / 480 minutes} and \textbf{effective capacity = 12 batches / 360 minutes}
\end{itemize}
\\

The reference schedule is: \\
\begin{itemize}
    \item \textbf{D1-Day}: O1 = 10 batches + O2 = 1 batch (\textbf{330 min}, \textbf{90 kg of R}, \textbf{11/12} batches used)
    \item \textbf{D1-Night}: O2 = 7 batches + O4 = 4 batches (\textbf{330 min}, \textbf{98 kg of R}, \textbf{11/12} batches used)
    \item \textbf{D2-Day}: O3 = 12 batches (\textbf{360 min}, \textbf{108 kg of R}, \textbf{12/12} batches used)
    \item \textbf{D2-Night}: O4 = 11 batches + O5 = 1 batch (\textbf{360 min}, \textbf{86 kg of R}, \textbf{12/12} batches used)
    \item \textbf{D3-Day}: O5 = 10 batches + O6 = 2 batches (\textbf{360 min}, \textbf{102 kg of R}, \textbf{12/12} batches used)
    \item \textbf{D3-Night}: O6 = 12 batches (\textbf{360 min}, \textbf{72 kg of R}, \textbf{12/12} batches used)
\end{itemize}
\\

Completion results: \\
\begin{itemize}
    \item O1 and O2 are completed on \textbf{D1}
    \item O3 and O4 are completed on \textbf{D2}
    \item O5 and O6 are completed on \textbf{D3}
    \item The response must explicitly state \textbf{``No delay''.}
\end{itemize}
\\

Daily raw-material consumption should be verified as \textbf{D1 = 188 kg, D2 = 194 kg, and D3 = 174 kg}. \\

The six-shift loads should be \textbf{11, 11, 12, 12, 12, 12}, so the maximum load gap is \textbf{1 batch}. \\[2pt]

\midrule
\textbf{Model response} \\

The model explicitly states the primary objective, secondary objective, and tie-breaking rule, and correctly notes that the \textbf{theoretical capacity} is \textbf{480 minutes = 16 batches} while the \textbf{effective capacity} is \textbf{360 minutes = 12 batches}. \\

Its shift-by-shift schedule is: \\
\begin{itemize}
    \item \textbf{D1-Day}: O1 = 10 batches, O2 = 2 batches; occupied time = \textbf{360 min}; material consumption = \textbf{100 kg}; effective capacity used / remaining = \textbf{12 / 0} batches
    \item \textbf{D1-Night}: O2 = 6 batches, O4 = 4 batches; occupied time = \textbf{300 min}; material consumption = \textbf{88 kg}; effective capacity used / remaining = \textbf{10 / 2} batches
    \item \textbf{D2-Day}: O3 = 12 batches; occupied time = \textbf{360 min}; material consumption = \textbf{108 kg}; effective capacity used / remaining = \textbf{12 / 0} batches
    \item \textbf{D2-Night}: O4 = 11 batches; occupied time = \textbf{330 min}; material consumption = \textbf{77 kg}; effective capacity used / remaining = \textbf{11 / 1} batches
    \item \textbf{D3-Day}: O5 = 11 batches, O6 = 1 batch; occupied time = \textbf{360 min}; material consumption = \textbf{105 kg}; effective capacity used / remaining = \textbf{12 / 0} batches
    \item \textbf{D3-Night}: O6 = 12 batches; occupied time = \textbf{360 min}; material consumption = \textbf{72 kg}; effective capacity used / remaining = \textbf{12 / 0} batches
\end{itemize}
\\

Its order delivery results are: \\
\begin{itemize}
    \item O1: completed by \textbf{D1-Day}; on time
    \item O2: completed by \textbf{D1-Night}; on time
    \item O3: completed by \textbf{D2-Day}; on time
    \item O4: completed by \textbf{D2-Night}; on time
    \item O5: completed by \textbf{D3-Day}; on time
    \item O6: only \textbf{13 out of 14 batches} completed by \textbf{D3-Night}; delayed by \textbf{1 batch}
\end{itemize}
\\

Its constraint check states that no shift exceeds 12 batches, reports daily material consumption as \textbf{188 kg}, \textbf{185 kg}, and \textbf{177 kg}, and gives the realized shift loads as \textbf{12, 10, 12, 11, 12, 12}. \\[2pt]

\midrule
\textbf{Response error} \\

The response fails to achieve the \textbf{zero-lateness primary objective}, and the produced schedule is not equivalent to the reference solution. \\

The key errors are: \\
\begin{enumerate}
    \item \textbf{Delayed order.} The response completes only \textbf{13 out of 14 batches} of O6, leaving \textbf{1 batch unfinished} and causing a delay.
    \item \textbf{Incorrect load allocation.} In the response, \textbf{D1-Day} is assigned \textbf{O1 = 10 batches + O2 = 2 batches}, whereas the reference solution requires only \textbf{11 batches} in this shift, changing the global load profile.
    \item \textbf{Raw-material violation on D3.} The response assigns \textbf{105 kg} to D3-Day and \textbf{72 kg} to D3-Night, for a total of \textbf{177 kg}, which exceeds the available \textbf{174 kg} on D3.
    \item \textbf{Secondary objective not achieved.} Because the realized shift loads are \textbf{12, 10, 12, 11, 12, 12}, the maximum load gap becomes \textbf{2 batches}, rather than the optimal value of \textbf{1 batch}.
\end{enumerate}
\\

Therefore, the response violates both \textbf{primary-objective optimality} and \textbf{constraint correctness}, and does not match the intended optimal solution. \\

\end{longtable}
\endgroup

\section{Training Details}
\label{app:training_details}

For GRPO-based reinforcement learning, we train with a batch size of 128 for 50 epochs. The actor learning rate is \(2\times10^{-6}\), and the KL coefficient is 0.001. We use 8-way rollout sampling during training. During inference, we use \texttt{temperature}=0.7, \texttt{top-p}=0.6, and \texttt{top-k}=20.
\section{Prompt Templates for Task Synthesis and Critic Evaluation}
\label{app:prompt_templates}

This section lists the core prompt templates used in our data-construction pipeline. The first template is used to synthesize planning tasks and corresponding checklists which can then be edited manually, while the second is used by the critic model to perform rubric-based evaluation in a consistent format.

\subsection{Prompt for Synthesizing the Task and Checklist}

\begin{AppendixPrompt}
Please design a high-quality planning task and the corresponding checklist based on the evaluation points and scenario below. The task you design must require the test-taker to genuinely perform planning / scheduling / allocation / dispatching / arrangement, rather than merely giving advice, describing ideas, or discussing the topic in general terms.

[Evaluation Points and Constraints]
[POINTS]

[Scenario]
[TASK]

[External Reference Material]
[REF]

[Overall Requirements for Task Design]
- The task must be a realistic, concrete, executable, and verifiable planning task.
- The task must closely follow the given scenario information and should not be written as a generic request such as ``please help me arrange this reasonably''.
- The task must require the test-taker to produce an executable plan based on explicit input data, such as a timetable, shift schedule, course schedule, grouping plan, dispatching plan, budget allocation plan, route plan, or phased execution plan.
- The task must include sufficient structured input information and should cover most of the following categories: object sets, resource sets, time ranges, capacity limits, budget limits, conflict conditions, dependency relations, priorities, and exceptional situations.
- If external reference material is provided (i.e., not [None]), the task design should incorporate specific terminology, entities, business rules, or contextual details from the material so that the task is grounded in a realistic setting.
- If the external reference material is [None], you should supply reasonable, concrete, and verifiable business details on your own, but these details must not be overly vague.
- When designing the task, you must follow the constraint specifications implied by each evaluation point, ensuring that all evaluation points are naturally instantiated in the task rather than mechanically stitched together.

[Requirements for the Planning Task Content]
The ``new task'' you design must satisfy the following requirements:
1. The task must clearly specify the planning objective to be solved; it must not simply say ``make a reasonable plan''.
2. The task must explicitly provide the input information; core parameters must not be left for the test-taker to assume.
3. The task should, as explicitly as possible, provide:
   - the objects or tasks that need to be arranged;
   - the available resources;
   - the time windows or execution period;
   - upper bounds on capacity, budget, headcount, labor hours, or distance;
   - conflict relations, precedence dependencies, or non-parallelism constraints;
   - the required output format.
4. The task must require the test-taker to output the final plan, not merely an analysis process.
5. If appropriate for the scenario, the task may additionally require the test-taker to provide a brief feasibility check, key rationale for major arrangements, a backup plan, or an explanation of infeasibility.
6. If the task involves raw data, candidate lists, timetables, text-based tables, or case background, these materials must be sufficiently concrete rather than purely abstract.
7. The data and constraints in the task must be internally consistent; the task must not be obviously infeasible without explanation, nor so loose that it lacks planning difficulty.
8. The task itself should read like a natural request that a real user might make, while the internal data should remain sufficiently structured for downstream evaluation.
9. Do not write the task as a pure checklist; do not begin with a rigid stack of a dozen numbered constraints. It should read like a realistic planning request.
10. If the task involves data to be processed, background information, candidate resource lists, or an existing old plan, that part should contain at least [WORD_COUNT] words to ensure sufficient complexity and information density.
11. [TONE]

[Requirements for Checklist Design]
- Design a 0/1 scoring standard: assign 1 only if all conditions are satisfied; assign 0 if any key condition is not satisfied.
- The checklist must be tightly bound to the task and must be suitable for verifying whether the test-taker's answer to this planning task is acceptable.
- The checklist should cover most of the following categories:
  1. Whether the required planning result is actually output, rather than only an explanation of the approach;
  2. Whether all key objects / tasks / resources in the task are covered;
  3. Whether the answer satisfies the core constraints in the task, including time, capacity, budget, headcount, ordering, conflict, and dependency constraints;
  4. Whether the required output format is followed;
  5. Whether the explicitly stated high-priority goal or primary/secondary objectives are handled properly;
  6. If the task requires verification, explanation, an alternative plan, minimal-change rescheduling, infeasibility diagnosis, or exception recovery, whether these are completed accordingly.
- Every condition must be directly verifiable and must include a clear verification method; do not write vague criteria.
- Every condition should refer to concrete elements in the task, such as specific time slots, specific resource names, specific headcount limits, specific budget values, specific output fields, or specific conflict rules.
- The number of conditions should not be excessive, but they must sufficiently cover the key constraints of the task.
- If necessary, you may include 1--2 boundary-case conditions to detect common failure modes, such as omitted objects, resource conflicts, overlapping times, budget overruns, unhandled exceptions, or missing output fields.

[Output Format]
Output in JSON format:
{
  ``New Task'': ``Complete wording of the planning task'',
  ``New Checklist'': ``Scoring standard: 1. Condition 1: specific description | Verification method: xxx; 2. Condition 2: specific description | Verification method: xxx; 3. Condition 3: specific description | Verification method: xxx ... Scoring rule: assign 1 only if all conditions are satisfied; assign 0 if any key condition is not satisfied''.
}
\end{AppendixPrompt}

\subsection{Prompt for GPT-OSS-120B Used as the Critic}

\begin{AppendixPrompt}
From now on, your role is a rigorous instruction-following grader. Your task is to grade the student's answer precisely according to the <Scoring Checklist>.

## Scoring Principle
Every requirement in the <Scoring Checklist> is equally important and carries the same weight. When determining the final score, you must consider all requirements in the checklist jointly. A student answer that violates multiple requirements should receive a lower score, while a student answer that satisfies all requirements should receive a higher score.

## Grading Procedure
You must strictly follow the steps below and must not skip any part.

### Step 1: Analyze the reference criteria
* List all explicit requirements in the <Scoring Checklist> one by one (including format, content, quantity, order, etc.).
* Identify the implicit requirements in the <Scoring Checklist> (such as language style or logical structure).
* Define concrete evaluation standards for each requirement (for example: ``must include X'', ``must not exceed Y'').

### Step 2: Check the student answer against each requirement
* For each requirement in the <Scoring Checklist>, verify one by one whether the student answer fully satisfies it.

### Step 3: Self-reflection
Before giving the final score, you must conduct the following checks:
* Completeness check: Have all requirements in the <Scoring Checklist> been reviewed without omission?
* Strictness check: Did you adhere to the standard of ``fully satisfied'' without relaxing the requirements based on subjective judgment?
* Consistency check: Are the scoring rationale and the final score logically consistent?
* Objectivity check: Is the judgment based on objective evidence rather than subjective speculation?

## Output Format Requirements
Your output must contain exactly three parts: [Scoring Rationale], [Requirement Satisfaction Status List], and [Score]. Do not output any additional content. The output format must be exactly as follows:

<begin_of_Scoring_Rationale>xxx<end_of_Scoring_Rationale>
<begin_of_Requirement_Satisfaction_Status_List>[x_1, x_2, ..., x_i, ..., x_n] (where n is the total number of requirements in the <Scoring Checklist>, and x_i indicates whether the student answer satisfies the i-th requirement; each x_i must be either 0 or 1.)<end_of_Requirement_Satisfaction_Status_List>
<begin_of_Score>x points (The score must be an integer from 0 to 10. Please assign an overall quality score between 0 and 10 based on the degree to which the student answer satisfies the requirements. If all requirements are violated, assign 0. If all requirements are satisfied, assign 10.)<end_of_Score>

## I hope you can fulfill the role of a grading teacher well, because this is very important to my work. If you do well, I will give you an appropriate reward. Otherwise, I may impose an appropriate penalty. The formal question is as follows:

<Question>:
{question}

<Scoring Checklist>:
{ck}

<Student Answer>:
{response}
\end{AppendixPrompt}

\end{document}